\documentclass[journal]{IEEEtran}
\IEEEoverridecommandlockouts
\usepackage{stfloats,color}
\usepackage{float}
\usepackage{amsfonts}
\usepackage[cmex10]{amsmath}
\usepackage{nth}
\usepackage{graphicx}
\usepackage{setspace}
\usepackage{ragged2e}
\usepackage{cite}
\usepackage{bbm}
\usepackage{colortbl}
\definecolor{maroon}{cmyk}{0,0.87,0.68,0.32}
\definecolor{Gray}{gray}{0.9}

\usepackage{filecontents}
\usepackage{array}
\usepackage{algpseudocode}
\usepackage{mdwmath}
\usepackage{multirow}
\usepackage{makecell}
\usepackage[caption=false, ...]{subfig}

\usepackage{url}
\usepackage{times}
\usepackage{diagbox}
\usepackage{epsfig}
\usepackage{latexsym}
\usepackage{epstopdf}
\usepackage{verbatim}
\usepackage{units}
\usepackage{amsthm}
\usepackage{mdwlist}
\graphicspath{{./figures/}}
\usepackage{esvect}
\usepackage[linesnumbered,algoruled,boxed,lined]{algorithm2e}
\usepackage{tabularx,colortbl}
\usepackage[unicode=true, linktocpage, linkbordercolor={0.5 0.5 1}, citebordercolor={0.5 1 0.5}, linkcolor=blue]{hyperref}

\usepackage{booktabs}
\usepackage{siunitx}

\usepackage[utf8x]{inputenc}
\usepackage[english]{babel}

\usepackage{lipsum}

\makeatletter
\def\footnoterule{\relax%
  \kern-5pt
  \hbox to \columnwidth{\hfill\vrule width 0.9\columnwidth height 0.4pt\hfill}
  \kern4.6pt}
\makeatother

\begin{document}

\title{Human Intracranial EEG Quantitative Analysis and Automatic Feature Learning for Epileptic Seizure Prediction}

\author{Ramy~Hussein$^{\star}$, Mohamed~Osama~Ahmed, Rabab~Ward, Z.~Jane~Wang, Levin~Kuhlmann, and~Yi~Guo
\thanks{The first author is funded by Vanier Canada Graduate Scholarship from the Natural Sciences and Engineering Research Council of Canada (NSERC).

$^\star$Corresponding Author: ramy@ece.ubc.ca 

Ramy Hussein, Z.~Jane~Wang and Rabab~Ward are with the Department of Electrical and Computer Engineering, University of British Columbia, Vancouver, BC V6T 1Z4, Canada.

Mohamed~Osama~Ahmed is with the Department of Computer Science, University of British Columbia, Vancouver, BC V6T 1Z4, Canada.

Levin~Kuhlmann is with the Faculty of Information Technology, Monash University, the Centre for Human Psychopharmacology, Swinburne University of Technology \& Departments of Medicine and Biomedical Engineering, University of Melbourne, Australia.

Yi~Guo is with the Department of Neurology, Shenzhen People's Hospital, Shenzhen, Guangdong, China.

}
}
\maketitle

\begin{abstract}
\textit{Objevtive:}
The aim of this study is to develop an efficient and reliable epileptic seizure prediction system using intracranial EEG (iEEG) data, especially for people with drug-resistant epilepsy. The prediction procedure should yield accurate results in a fast enough fashion to alert patients of impending seizures.
\textit{Methods:}
We quantitatively analyze the human iEEG data to obtain insights into how the human brain behaves before and between epileptic seizures. We then introduce an efficient pre-processing method for reducing the data size and converting the time-series iEEG data into an image-like format that can be used as inputs to convolutional neural networks (CNNs). Further, we propose a seizure prediction algorithm that uses cooperative multi-scale CNNs for automatic feature learning of iEEG data.
\textit{Results:} 
1) iEEG channels contain complementary information and excluding individual channels is not advisable to retain the spatial information needed for accurate prediction of epileptic seizures. 
2) The traditional PCA is not a reliable method for iEEG data reduction in seizure prediction.  
3) Hand-crafted iEEG features may not be suitable for reliable seizure prediction performance as the iEEG data varies between patients and over time for the same patient.
4) Seizure prediction results show that our algorithm outperforms existing methods by achieving an average sensitivity of 87.85\% and AUC score of 0.84. 
\textit{Conclusion:}
Understanding how the human brain behaves before seizure attacks and far from them facilitates better designs of epileptic seizure predictors.
\textit{Significance:}
Accurate seizure prediction algorithms can warn patients about the next seizure attack so they could avoid dangerous activities. Medications could then be administered to abort the impending seizure and minimize the risk of injury. Closed-loop seizure intervention systems could also help to prevent seizures in patients with drug-resistant epilepsy. 
\end{abstract}


\IEEEpeerreviewmaketitle

\section{Introduction}
\label{Section1}

\IEEEPARstart{E}{pilepsy} is a neurological disorder that affects around 70 million people worldwide \cite{Epilepsy}. It is characterized by recurrent seizures that strike without warning. Symptoms may range from brief suspension of awareness to violent convulsions and sometimes loss of consciousness \cite{Symptoms}. Currently, anti-epileptic drugs are given to epileptic patients in sufficiently high dosages. These drugs could result in undesirable side effects such as tiredness, stomach discomfort, dizziness, or blurred vision. Unfortunately, 20-40\% of people with epilepsy continue to have seizures despite treatment \cite{french2007refractory}. Further, the patient’s quality of life is significantly degraded by the anxiety associated with the unpredictable nature of seizures and the consequences therefrom. This motivated researchers to develop seizure prediction systems \cite{d2003epileptic}. The ability to predict seizures with high accuracies would make individualized epilepsy treatment possible (\textit{e.g.}, tailored therapies with less side effects). With warnings of impending seizures, the patients can take their precautions and avoid any probable injuries. This vision inspired the proposed research.


Epilepsy surgery may improve the quality of life of epileptic patients who are drug-resistant to epilepsy. Brain surgery may reduce the number of seizure attacks and hence limit the risk of permanent brain damage. It may, however, involves serious risks such as stroke, paralysis, speech issues, memory problems, loss of vision, loss of motor skills, and sometimes more seizures \cite{tellez2010surgical}. Besides epilepsy surgery, people with drug-resistant epilepsy can benefit from methods that predict seizures far enough in advance. Despite the fact that epileptic seizures seem unpredictable and often occur without warning, recent investigations have demonstrated that seizures do not strike at random \cite{gadhoumi2016seizure, assi2017towards}. Intracranial electroencephalography (iEEG) is the prime tool used for forecasting epileptic seizure attacks. The iEEG data recorded preceding seizures are analyzed to identify pattern(s) that indicate upcoming seizures.


Over the past decade, researchers have made several attempts to develop robust iEEG-based seizure prediction methods. The prediction performance, however, was affected by the lack of long-term preictal (before seizures) and interictal (between seizuress, \textit{i.e.}, baseline) iEEG recordings \cite{mormann2006seizure, kuhlmann2010patient}. In 2013, Cook \textit{et al.} designed a seizure advisory system that can predict the probability of seizure occurrence in the minutes or hours in advance \cite{cook2013prediction}. Fifteen adults with refractory epilepsy were each implanted with 16 electrodes. This 16-channel seizure prediction device allows uninterrupted recording of intracranial EEG for a long period of time (6-36 months). The seizure prediction algorithm however yielded prediction performances with a large subject variability for 15 patients. For example, the highest seizure prediction sensitivity of 100\% was reached for two patients, while a low sensitivity of 17-45\% was reported for three patients. Improving prediction performances for these three patients is important to ensure that seizure prediction is possible for different patients with drug-resistant epilepsy. 


Recently, a big dataset of long-term iEEG recordings has been collected from the aforementioned three patients over 374-559 days. A subset of this dataset was made publicly available and used in the Melbourne University AES/MathWorks/NIH Seizure Prediction Competition organized in November, 2016 on Kaggle.com\footnote[1]{\href{https://www.kaggle.com/c/melbourne-university-seizure-prediction}{www.kaggle.com/c/melbourne-university-seizure-prediction}}. Kuhlmann \textit{et al.} describe the human iEEG dataset used in this contest and the results achieved by the top eight seizure prediction algorithms \cite{kuhlmann2018epilepsyecosystem}. These algorithms adopted a diverse range of iEEG feature extraction, selection, and classification techniques; and achieved superior performance compared to those in \cite{cook2013prediction}. For example, the winning solution was based on an ensemble of 11 different machine learning models; each was designed subject-specific. A detailed description of the 11 models - including the extracted iEEG features as well as the utilized classification models - is provided in \cite{WinningSolution}. To assess the predictive power of these models, the area under the ROC curve was obtained for each model individually. The best reported prediction score for the contest test set was 0.854 \cite{kuhlmann2018epilepsyecosystem, WinningSolution}.


In this work, we propose a Convolutional Neural Network (CNN)-based algorithm for reliable prediction of epileptic seizures. Major contributions of this study are as follows: 
(1) We first provide an extensive quantitative analysis of the human iEEG data preceding and between epileptic seizures; 
(2) We introduce an efficient pre-processing strategy for transferring time-series iEEG data into image-like format that can better leverage the power of CNNs;
(3) We propose a multi-scale CNN architecture that can learn different representations of iEEG data efficiently; and
(4) We demonstrate that the proposed seizure prediction algorithm works reliably for different patients with refractory focal epilepsy, and outperforms the state-of-the-art algorithms for this problem.
To the best of our knowledge, this is the first attempt to automatically learn both local and high-abstracted iEEG features simultaneously. This is in contrast to previous studies that rely on extracting hand-crafted features.

\section{Human iEEG Quantitative Analysis}
\label{Section2}

In this section, we first describe the human intracranial EEG data used in our study and then present some hidden insights of the data, in order to better understand the unpredictable nature of epileptic seizures that we have been attempting to quantify.


\subsection{Subjects and Data}

The iEEG data used is from the 2016 Kaggle seizure prediction competition and is described in \cite{kuhlmann2018epilepsyecosystem}. The data were recorded constantly from humans suffering from refractory (drug-resistant) focal epilepsy using the NeuroVista Seizure Advisory System (described in \cite{cook2013prediction}). Sixteen electrodes (4$\times$4 contact strips) were implanted in every patient, directed to the presumed seizure focus, and connected to a telemetry unit embedded in the subclavicular area. A rechargeable battery powered the embedded device. Data were sampled at 400Hz, digitized using a 16-bit analog-to-digital converter, wirelessly transmitted to an external hand-held advisory device, and continuously saved in a removable flash drive.


Three drug-resistant patients who also had the least seizure prediction performance in \cite{cook2013prediction} were chosen for our study. The selection of these patients was propelled by the intention to demonstrate that accurate seizure prediction is also feasible for patients with drug-resistant epilepsy. The large number of seizures recorded per patient ($\sim$380) gives us the chance to explore the common signature within the iEEG preceding seizures of every patient. The three patients were females and they all had resective surgery before the trial. The first, second, and third patient were 22, 51, and 53 years old at the time of the trial, but were diagnosed with epilepsy at age of 16, 10, and 15, respectively. The first patient suffered from parietotemporal focal epilepsy and was receiving the antiepileptic drugs of carbamazepine, lamotrigine, and phenytoin. The second patient was diagnosed with occipitoparietal focal epilepsy and was taking carbamazepine medication. The third patient had seizures of frontotemporal origin and was taking perampanel, phenytoin, and Lacosamide for treatment.


Neuroscientists have found that the temporal dynamics of the brain activity of epileptic patients can often be categorized into four different states: \textit{preictal} (prior to seizure), \textit{ictal} (seizure), \textit{postictal} (after seizure), and \textit{interictal} (between seizures). The key challenge in seizure prediction is to differentiate between the \textit{preictal} and \textit{interictal} brain states in humans with epilepsy. Only lead seizures, characterized as seizures occurring at least 4 hours after a previous seizure, were used for every patient. The captured iEEG data were labeled and separated into training and testing sets (for training and testing the proposed seizure prediction method). To mimic real-life clinical settings, the testing data was recorded quite much later after recording the training data. To avert the signal non-stationarity in the immediate time period following sensors' implantation, both sets of data (training and testing) were obtained from data recorded between 1 and 7 months after sensors' implantation.

\begin{figure}[!t]
	\centering
	\includegraphics[width=1.0\columnwidth, height=0.20\columnwidth]{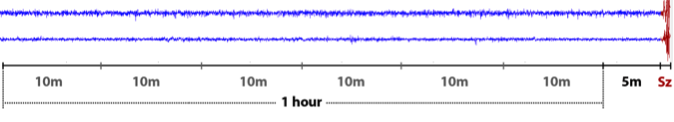}
	\caption{Examples of one-hour preictal iEEG clips collected by a 16-channel device with a 5-minute offset before seizures. For convenience, only two channels are plotted.}
	\label{Fig_PreictalEEG}
\end{figure}


As shown in Fig.~\ref{Fig_PreictalEEG}, 65-min preictal data clips were extracted preceding the lead seizures. They are divided into six 10-minute clips followed by a 5-minute offset segment before each seizure. Every two consecutive segments were also separated by a 10-second gap. In a similar manner, interictal clips were segmented from 61 minutes long recordings that started at arbitrarily time with a minimum gap of 3 hours before and 4 hours after any seizure. They also were split into six 10-minute clips with 10-second spacing. A summary of the data for every patient and the number of iEEG clips in the training and testing data sets are shown in Table~\ref{Tab_iEEGDescription}.


\begin{table*}[!t]
	\centering
	\caption{Description of the seizure prediction intracranial EEG dataset \cite{cook2013prediction}.}
	\label{Tab_iEEGDescription}
	\begin{tabular}{c c c c c c c c c c c}
		\hline
		\hline
		Patient & Age & Gender & Recording duration & Seizures & Lead seizures & \multicolumn{2}{c}{No. of training clips} & \multicolumn{2}{c}{No. of testing clips}  \\
		& (years) & & (days) & &  & Preictal & Interictal &  Preictal & Interictal  \\
		\hline
		1 & 22 & Female & 559 & 390 & 231 & 256 & 570 & 16 & 46 \\
		2 & 51 & Female & 393 & 204 & 186 & 222 & 1836 & 18 & 279 \\
		3 & 53 & Female & 374 & 545 & 216 & 255 & 1908 & 18 & 188 \\
		\hline
		\hline
	\end{tabular}
\end{table*}

\subsection{Are Interictal and Preictal iEEGs Statistically Different?}

\begin{figure}[!t]
	\centering
	\includegraphics[width=1.0\columnwidth, height=0.70\columnwidth]{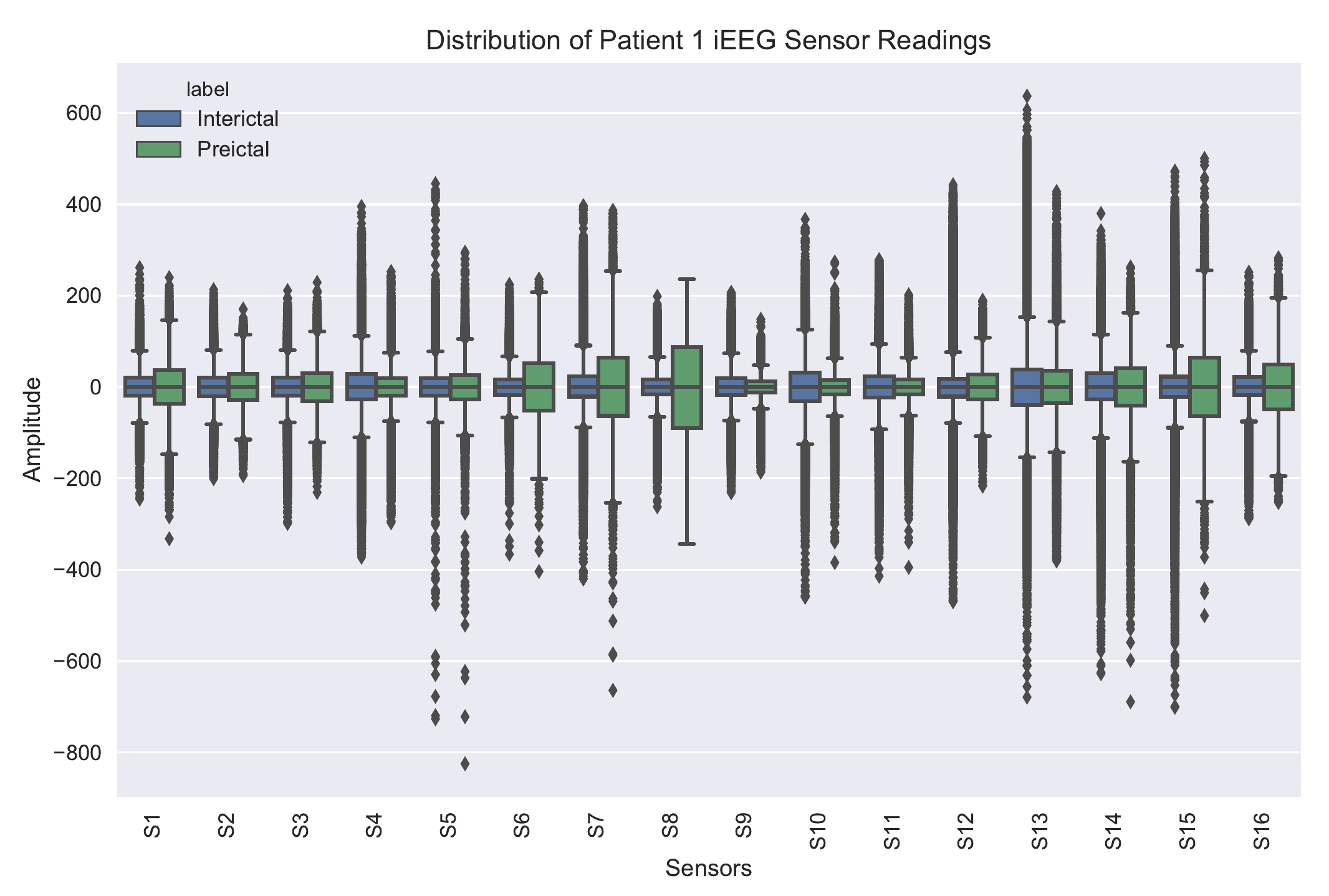}
	\caption{Boxplots of Patient~1’s \textit{interictal} and \textit{preictal} iEEG data.}
	\label{Fig_BoxPlot_Patient1}
\end{figure}

Most seizure prediction problems focus on the discrimination between the interictal (baseline) and preictal brain states. The question that arises here is: Are the interictal and preictal iEEGs statistically different? If yes, can we use their statistical features to differentiate between them? In an effort to address this question, we use descriptive statistics to describe whether these data are similar or different. The boxplot is a descriptive statistics tool that presents the data distribution using the five number summary: minimum, first quartile, median, third quartile, and maximum. To interpret a boxplot, we focus on three measures:


\begin{itemize}
	\item Median: The median is indicated by the horizontal line in the center of the box.
	\item Range: The range describes the spread of the data, represented by the vertical distance between the minimum and maximum values. 
	\item Interquartile Range (IQR): IQR shows the likely range of variations and is represented by the length of the box spanning the first quartile to the third quartile.
\end{itemize}


Fig.~\ref{Fig_BoxPlot_Patient1} shows a comparative boxplot for Patient~1’s interictal and preictal iEEG recordings. It can be noticed that, for all 16 sensors, the median values of the interictal and the preictal iEEG are both around zero. Moreover, for most of the sensor readings ($\sim$11 sensors), the overall range and interquartile range of the preictal data are much greater than those of the interictal data. However, for the rest of the sensor readings, the overall range and interquartile range of the preictal data are smaller than those of the interictal data. Interestingly, both data sets display potential outliers at both ends.

\begin{figure}[!t]
	\centering
	\includegraphics[width=1.0\columnwidth, height=0.70\columnwidth]{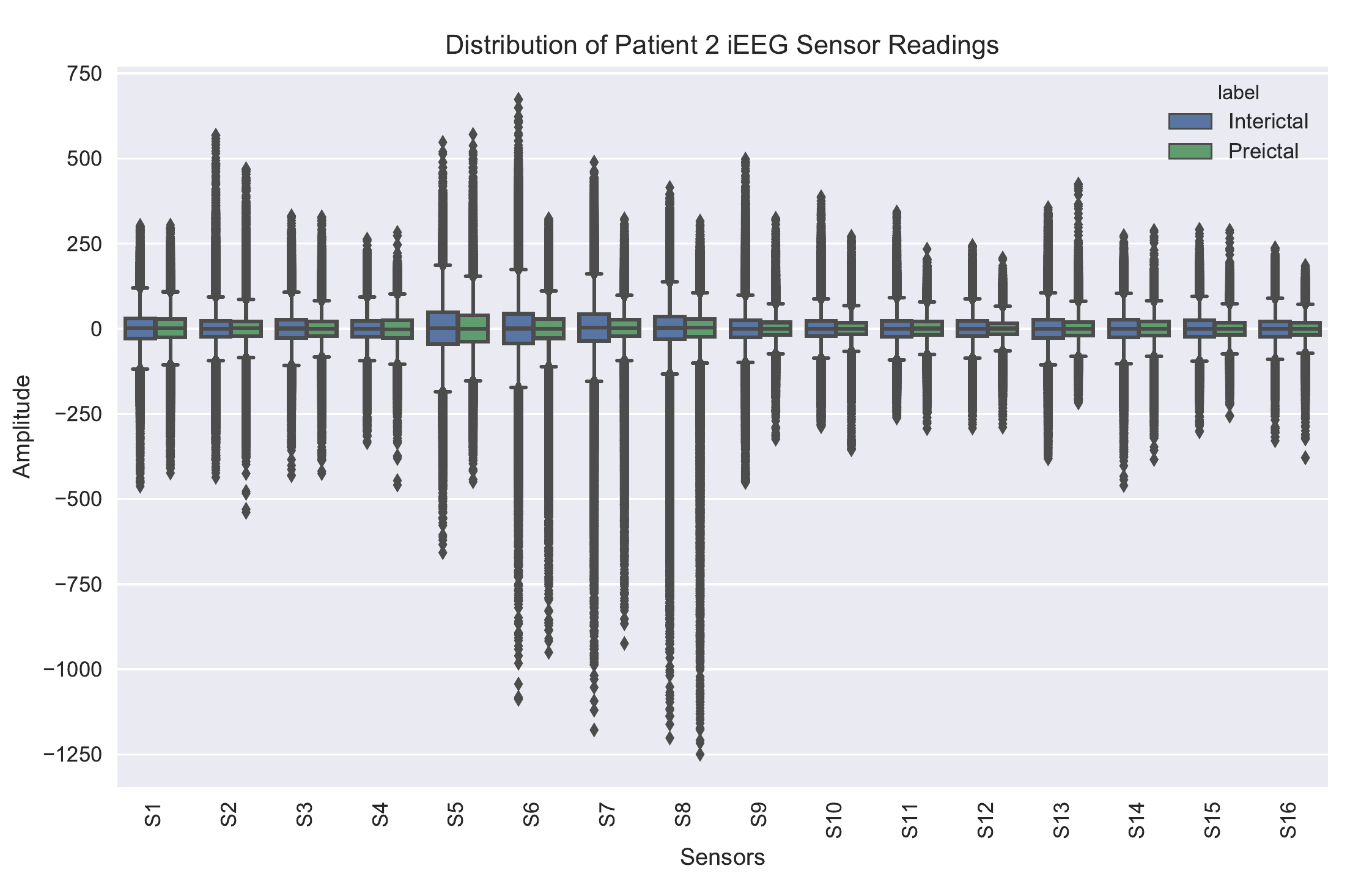}
	\caption{Boxplots of Patient~2’s \textit{interictal} and \textit{preictal} iEEG data.}
	\label{Fig_BoxPlot_Patient2}
\end{figure}

Despite the fact that the interictal and preictal data seem to have different dispersion (which implies that they are statistically different), we believe that such statistical features can only be used for patient-specific seizure prediction systems. Figure~\ref{Fig_BoxPlot_Patient2}, for instance, depicts that the distributions of interictal and preictal iEEG sensor readings for Patient~2 are quite different from those of Patient~1. Also, within the same iEEG data class (\textit{e.g.}, preictal), the overall range and IQR of the iEEG sensor readings of Patient~2 are less than those of Patient~1. Based on our observations for different patients, there is no typical trend in either interictal or preictal data across different epileptic patients. The iEEG data of each patient has its own characteristics and its statistical features can solely be meaningful for building a seizure prediction system for this particular patient. 


\subsection{Can PCA be Efficient for iEEG Data Reduction?}

Principal Component Analysis (PCA), is a prime tool for data reduction as it projects higher dimensional data to lower dimensional data. PCA has been widely used in reducing the dimensionality of the scalp (surface) EEG data, particularly for seizure onset detection. It also helps reduce the dimension of multi-channel EEG signals, and hence speed up the subsequent machine learning algorithms (\textit{e.g.}, feature extraction and classification), minimizing the computation time of the overall EEG-based diagnosis system of neurological disorders. PCA also helps remove the redundancy (correlation) in multi-variate EEG data, while maintaining most of the variance (information) in the observed variables. A useful measure is the ``explained variance'', which can be calculated from the eigenvalues to measure how much information can be represented by each of the principal components.  


The work presented in \cite{wang2017automatic} demonstrates how PCA can be effectively used in choosing the optimal feature subset from the original EEG feature set, and therefore improve the epileptic seizure detection performance. PCA has shown to be a valuable data reduction tool that preserves a significant amount of the EEG data variance and achieves stable seizure detection results. The question that arises here is: Can we use PCA to efficiently reduce the iEEG data dimensionality for epileptic seizure prediction? The iEEG dataset under study has 16 feature columns (16 sensors), a training set of 5,047 instances, and a test set of 565 instances (Table~\ref{Tab_iEEGDescription} shows the details of the used dataset). To answer the above question, we applied PCA to the interictal and preictal iEEG data under study. We then tested whether the 16-sensor readings can be mapped into a fewer number of principal components as follows: 


\begin{figure}[!t]
	\centering
	\includegraphics[width=1.0\columnwidth, height=0.75\columnwidth]{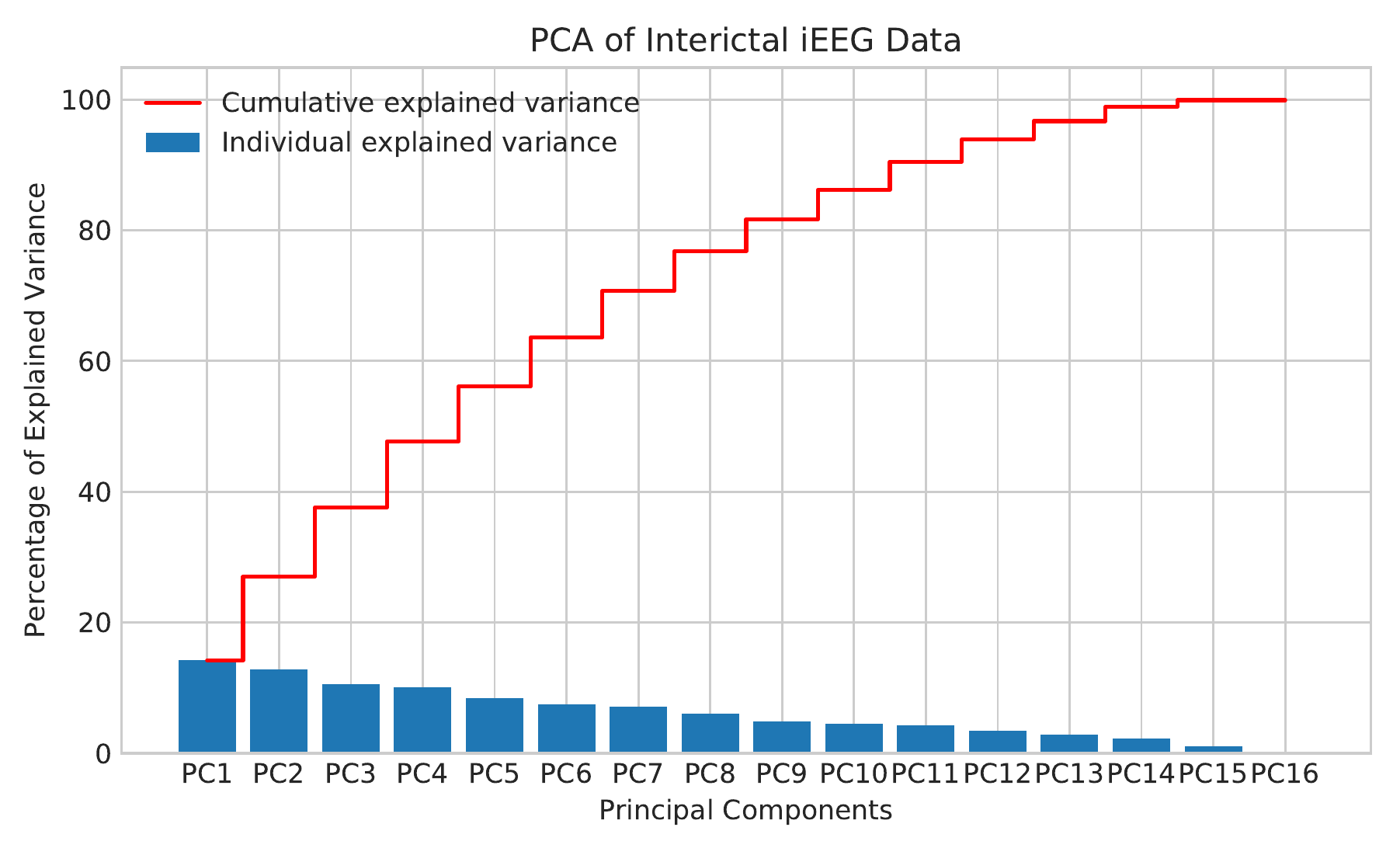}
	\caption{Explained variance by different principle components of interictal iEEG data for Patient~1.}
	\label{Fig_PCA_interictal}
\end{figure}

\begin{figure}[!t]
	\centering
	\includegraphics[width=1.0\columnwidth, height=0.75\columnwidth]{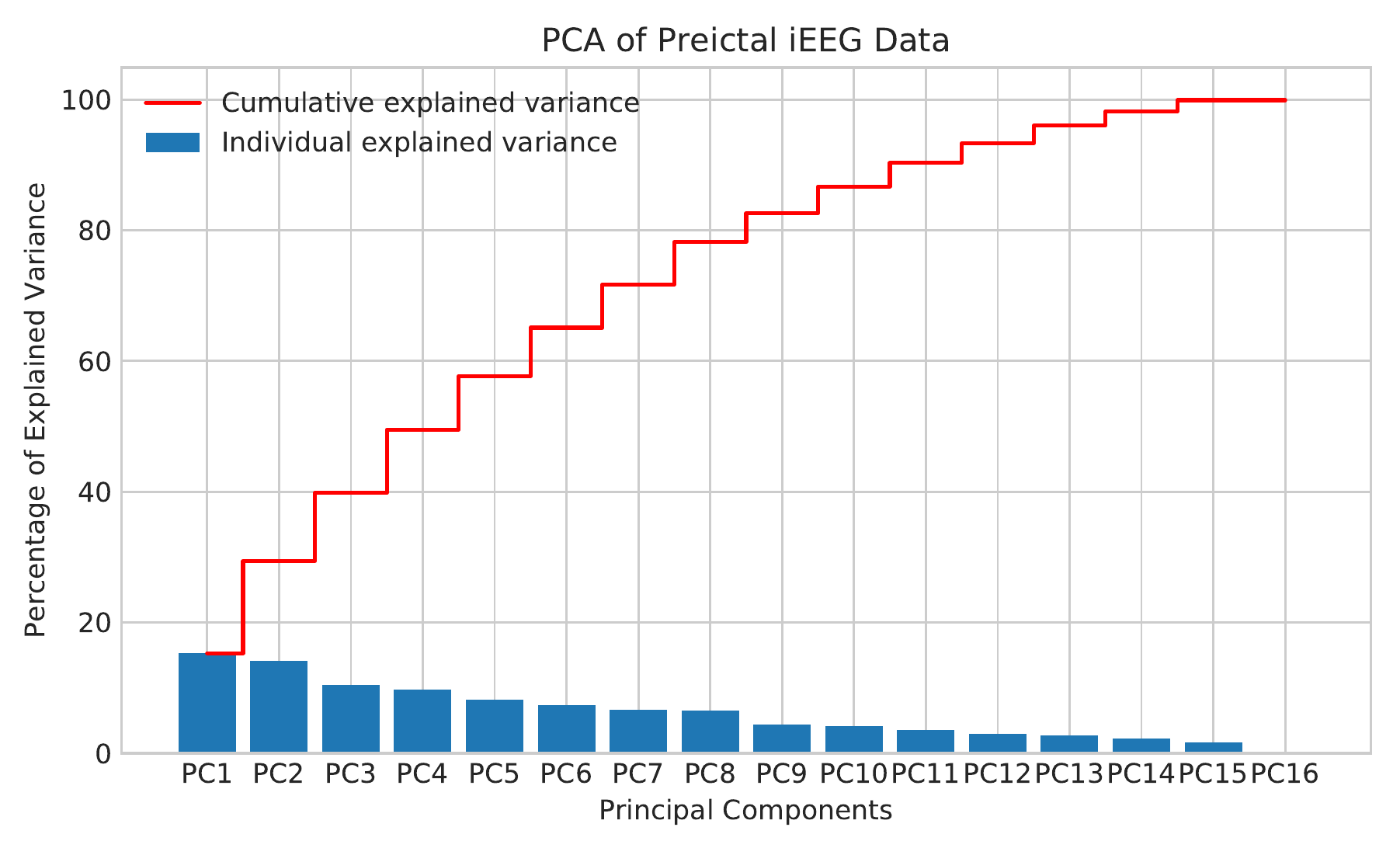}
	\caption{Explained variance by different principle components of preictal iEEG data for Patient~1.}
	\label{Fig_PCA_preictal}
\end{figure}

We chose the minimum number of principal components such that 95\% of the variance was retained. Figures~\ref{Fig_PCA_interictal} and \ref{Fig_PCA_preictal} show the individual principal components (colored in blue) and the cumulative principal components (colored in red) for both the interictal and preictal iEEG data clips for Patient~1. Unexpectedly, Fig.~\ref{Fig_PCA_interictal} shows that, for interictal iEEG, the first principal component accounts for a small amount of the data variance ($\sim$16\%), and 95\% of the variance is contained in 14 principal components. Only the 15th and 16th principal components can be dropped without losing too much information. Similarly, Fig.~\ref{Fig_PCA_preictal} represents the explained variance by 16 principal components of the preictal iEEG data. It provides useful insights on the distribution of variance across different principal components and suggests that, at least, 14 principal components are needed to preserve most of the data variance. This suggests that PCA is not a reliable method for iEEG data reduction for our problem. 


\subsection{Can We Exclude some EEG Channels/Sensors?}

Since PCA has failed to reduce the dimensionality of the human iEEG data when most of its variance is retained, researchers have considered turning some of the sensors off in order to reduce the amount of data and the computation time. The excluded channels are usually less relevant or redundant. It is a trade-off between selecting fewer EEG channels and retaining as much spatial information as possible. The work presented in \cite{shah2017optimizing} investigates the epileptic seizure detection performance for different channel configurations. As expected, using all EEG channels from a 10-20 EEG configuration achieves the best seizure detection performance. Choosing a moderately fewer number of EEG channels (\textit{e.g.}, 16 and 8) reduces the amount of data to 36-72\% of its original size, and results in a minor degradation in the seizure detection performance \cite{shah2017optimizing}.  

\begin{figure}[!t]
	\centering
	\includegraphics[width=1.0\columnwidth, height=0.75\columnwidth]{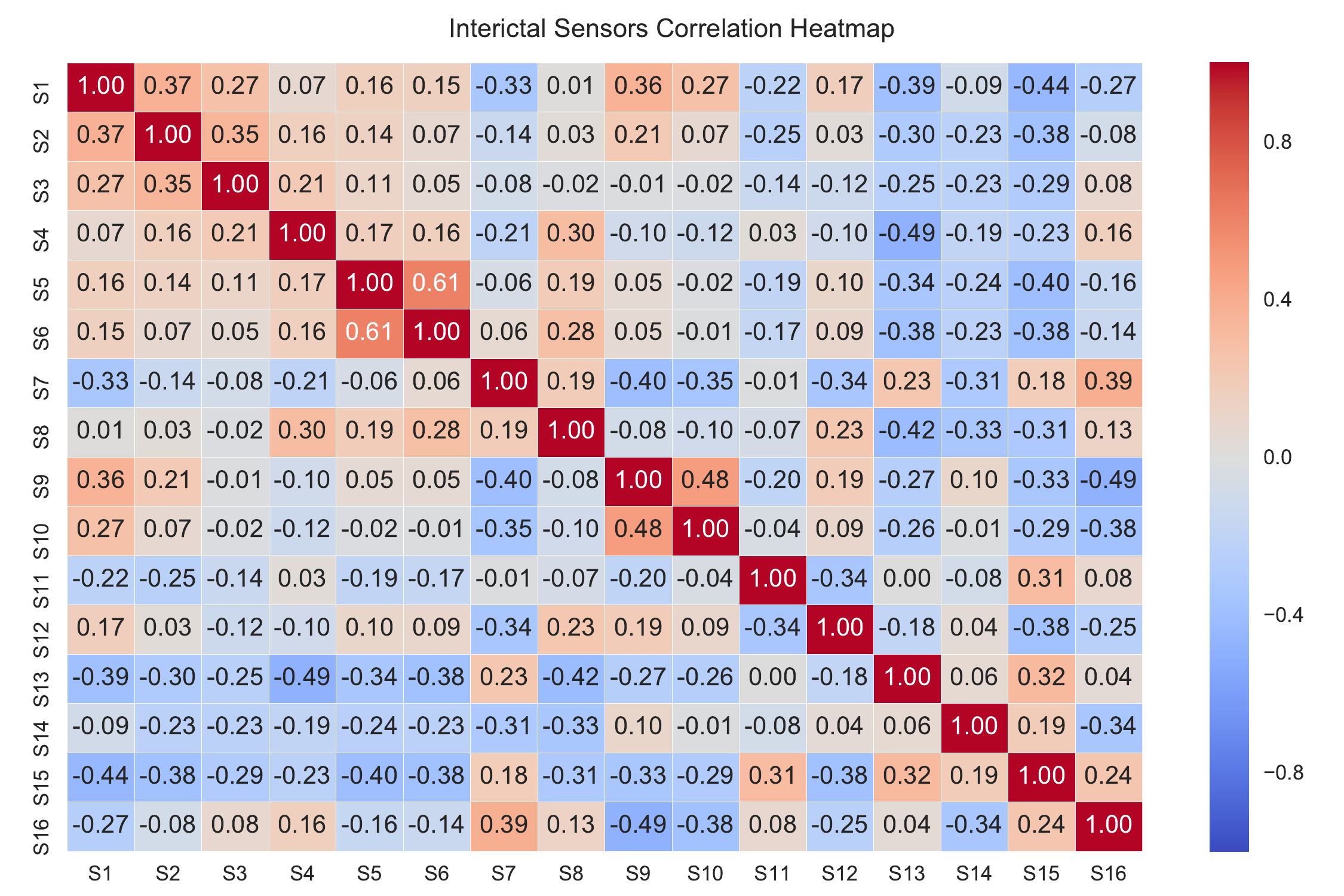}
	\caption{Heatmap of pairwise correlation values of Patient~1’s interictal iEEG's 16 sensors.}
	\label{Fig_Heatmap_interictal}
\end{figure}
\begin{figure}[!t]
	\centering
	\includegraphics[width=1.0\columnwidth, height=0.75\columnwidth]{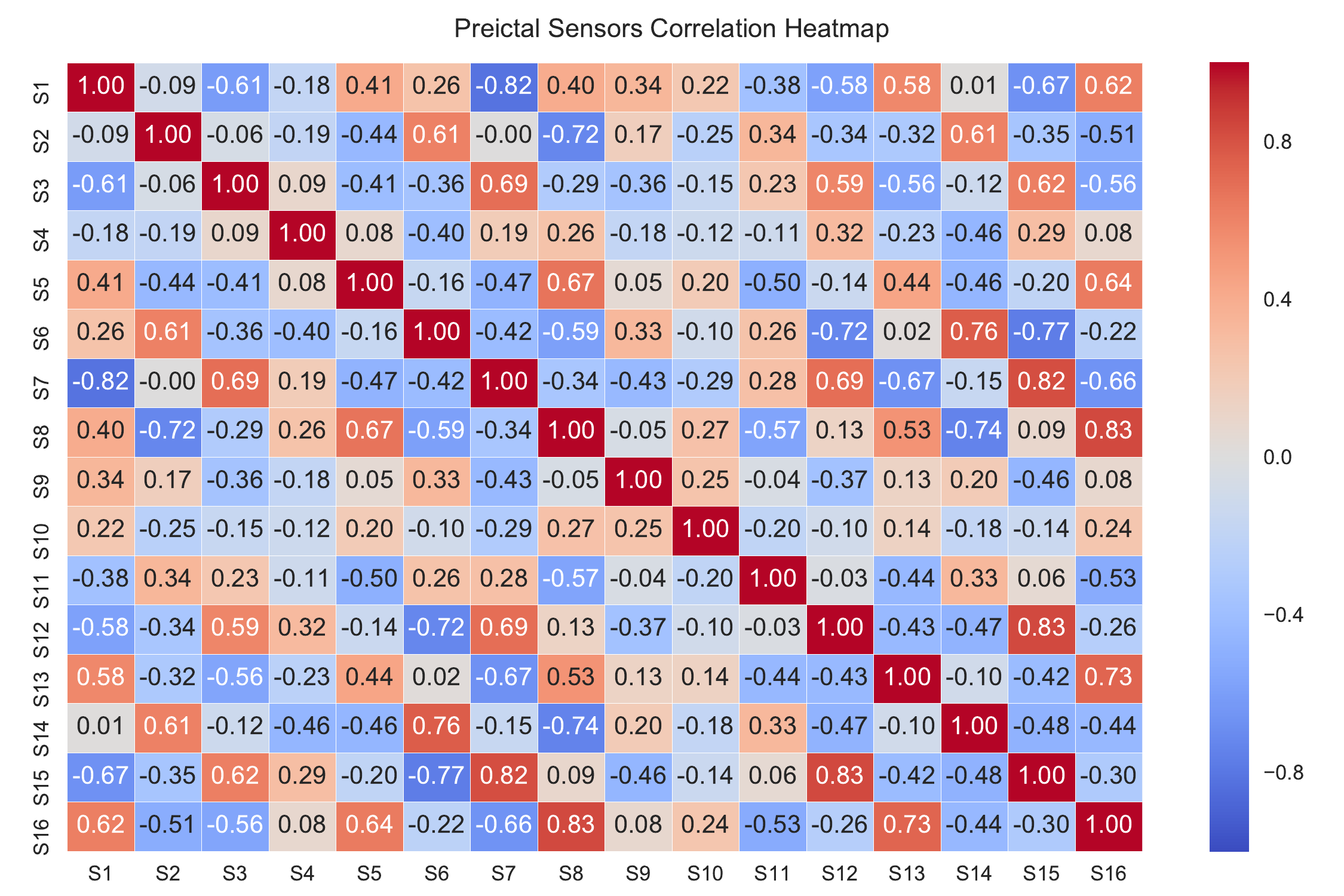}
	\caption{Heatmap of pairwise correlation values of Patient~1’s preictal iEEG's 16 sensors.}
	\label{Fig_Heatmap_preictal}
\end{figure}

Whilst many researchers have successfully used channel selection for surface EEG data reduction applied to seizure detection, there is no clear consensus regarding turning off any or some intracranial EEG channels in the seizure prediction task. The correlation heatmap gives us a way to gain some insights into what sensors (channels) are correlated and whether we can exclude certain sensors. Figures~\ref{Fig_Heatmap_interictal} and \ref{Fig_Heatmap_preictal} depict the correlation heatmaps of iEEG sensor data during the interictal and preictal brain states for Patient~1. The positive correlations are displayed by increasing the red hues and the negative correlations are displayed by increasing the blue shades. Faded red and blue colors denote low positive and negative correlations, respectively. The correlation coefficients are also included in the heatmaps, displaying the dependency relationships between 16 iEEG sensors.


In \cite{hall1999correlation}, Hall \textit{et al.} proposed a correlation-based feature selection method and concluded that: \textit{``a representative feature (sensor) subset is one that comprises features (sensors) highly correlated with the target (iEEG class), yet uncorrelated with each other''}. Following the same rule, we watched out for the iEEG sensors that are highly correlated during the interictal and preictal brain states. If a sensor’s predictive ability is covered by another, then it can safely be removed. As shown in Figures ~\ref{Fig_Heatmap_interictal} and \ref{Fig_Heatmap_preictal}, iEEG sensors are not heavily correlated neither in the interictal nor in the preictal brain states. The cross-correlation values range between $-$0.49 and 0.61 for the interictal iEEG sensor readings, and between $-$0.82 and 0.83 for the preictal iEEG sensor readings. Therefore, it is advised not to exclude any of the iEEG channels in order to retain the spatial information needed for accurate prediction of epileptic seizures.


\subsection{Are Neighbor iEEG Sensors more Correlated than Distant Ones?}

\begin{figure}[!t]
	\centering
	\includegraphics[width=1.0\columnwidth, height=0.40\columnwidth]{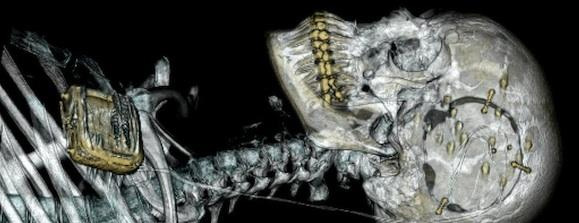}
	\caption{CT scan of the NeuroVista seizure advisory system implanted in a patient \cite{cook2013prediction}.}
	\label{Fig_CTscan}
\end{figure}

Fig.~\ref{Fig_CTscan} shows an example of Computerized Tomography (CT) scan of the head of one patient, and reveals the locations of 16 iEEG electrodes on the cortical surface of the brain. The questions here are: Are adjacent iEEG sensors more correlated than distant ones? If yes, are they positively or negatively correlated? In an attempt to answer these questions, we first identified the approximate location of the iEEG electrodes on the cerebral cortex of epileptic patients under study. Unlike most seizure prediction systems that deploy iEEG electrodes implanted across both brain hemispheres, the electrodes of the seizure advisory system used in the studied dataset were placed on one side over the small brain region presumed to include the epileptogenic zone. In patients with bilateral temporal lobe epilepsy, electrodes were positioned over the brain hemisphere that generates most seizures.


\begin{figure}[!t]
	\centering
	\includegraphics[width=1.0\columnwidth, height=0.80\columnwidth]{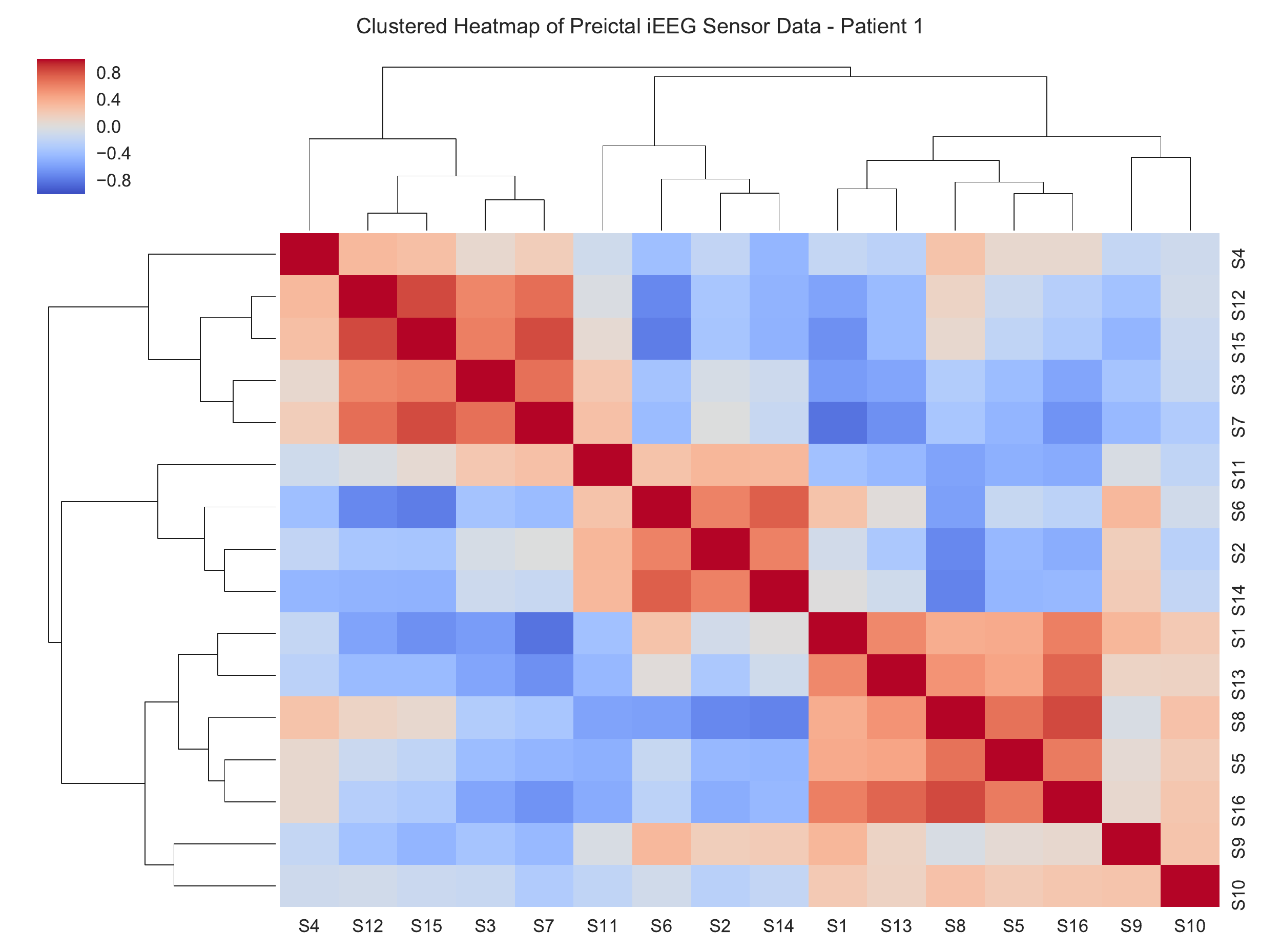}
	\caption{Hierarchically clustered preictal iEEG sensors with dendrograms and clusters in Patient~1.}
	\label{Fig_ClusteredHeatmap_Preictal_P1}
\end{figure}

\begin{figure}[!t]
	\centering
	\includegraphics[width=1.0\columnwidth, height=0.80\columnwidth]{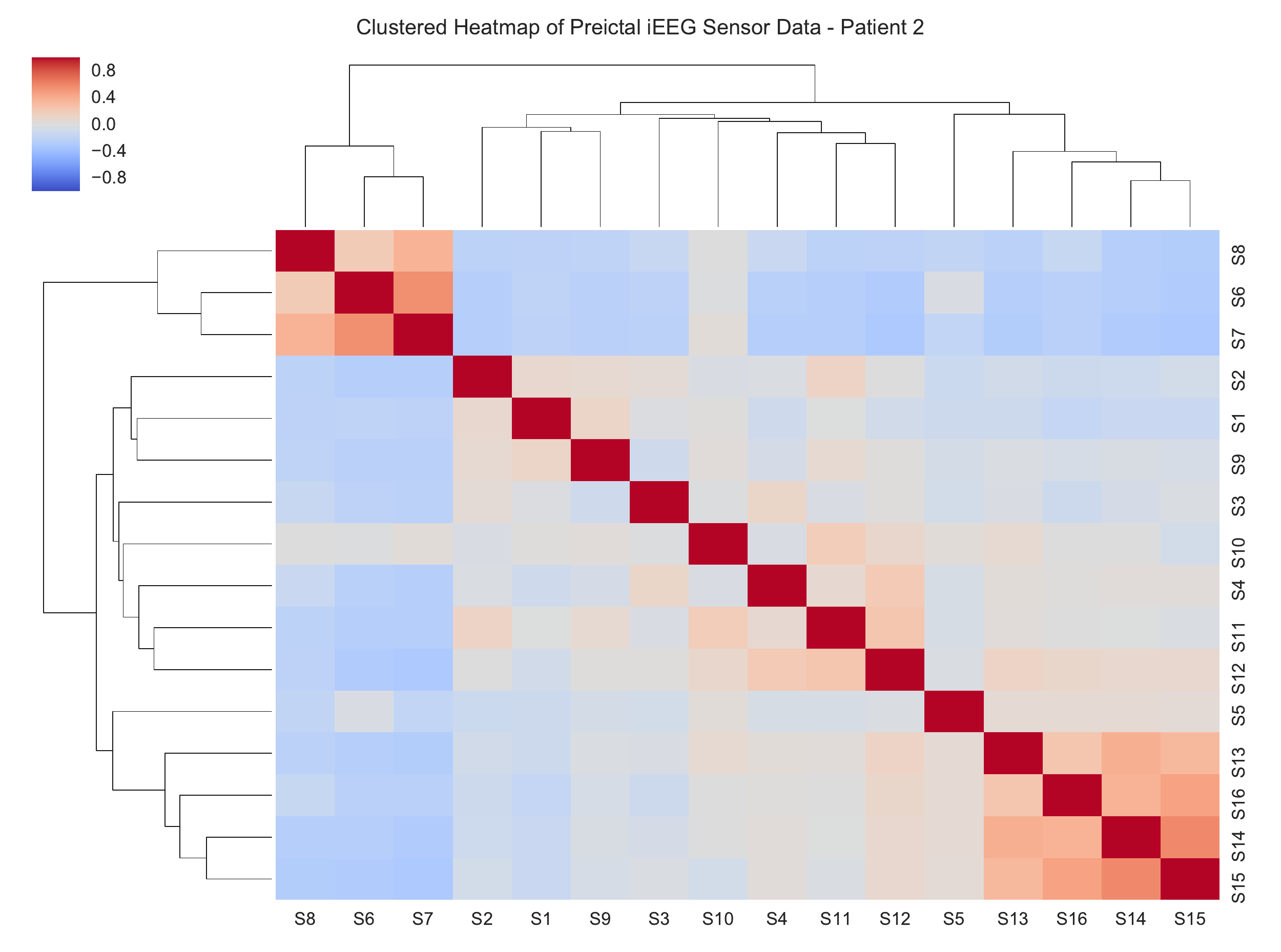}
	\caption{Hierarchically clustered preictal iEEG sensors with dendrograms and clusters in Patient~2.}
	\label{Fig_ClusteredHeatmap_Preictal_P2}
\end{figure}

Figures~\ref{Fig_ClusteredHeatmap_Preictal_P1} and \ref{Fig_ClusteredHeatmap_Preictal_P2} display the clustered heatmaps of preictal iEEG sensor data for Patients~1 and 2, respectively. As shown in Fig.~\ref{Fig_ClusteredHeatmap_Preictal_P1}, the iEEG sensor data of Patient~1 are grouped into three main clusters; each includes a set of sensors that have similar correlations. The first cluster comprises the sensors S3, S4, S7, S12, and S15 (S3 and S4 are neighbors), while the second cluster includes S2, S6, S11, and S14, and the third cluster includes S1, S5, S8, S9, S10, S13, and S16 (S9 and S10 are neighbors). As can be seen, adjacent sensors do not necessarily have analogous correlation profiles. These clustering methods may allow researchers to determine which particular sensors can act as seizure activity sources and which can act as sinks. However, categorizing iEEG sensors into ictal activity generators and sinks based on the correlation profiles is not well-established yet. In Fig.~\ref{Fig_ClusteredHeatmap_Preictal_P2} (for Patient~2), 
the first cluster includes S6, S7, and S8 (they all are neighbors), the second cluster includes S1, S2, S3, S4, S9, S10, S11, and S12 (where S1, S2, S3, and S4 are neighbors \& S9, S10, S11, and S12 are also neighbors), and the third cluster comprises S5, S13, S14, S15, and S16 (S5 is a non-neighbor to others). 


Despite the fact that both Patients~1 and 2 are diagnosed with focal epilepsy, it is important to realize that they have different preictal iEEG data clusters. Similarly, Patient~3 has a totally dissimilar iEEG clustered heatmap (omitted for the lack of space). The main reason why patients have different clustered heatmaps of preictal iEEGs is that they have different epileptogenic zones. Patient~1, for example, was diagnosed by parietal-temporal lobe epilepsy, while Patient~2 was diagnosed by occipito-parietal lobe epilepsy. In brief, the correlations between iEEG sensors vary depending on which brain region is affected and whether the seizure is focal (partial) or generalized.


To summarize, the conducted quantitative analysis of iEEG data suggests that iEEG data analysis for seizure prediction can be quite different from regular EEG data analysis for other biomedical classification/prediction problems; and it will be desirable to design an iEEG-based seizure prediction algorithm that (i) uses all iEEG channels for proficient feature extraction, (ii) avoids using PCA for iEEG dimensionality reduction, and (iii) avoids using domain-based hand-crafted features for iEEG classification. Such observations help explain why Epileptic Seizure Prediction using iEEG data remains a challenging topic and traditional machine learning algorithms using hand-crafted features did not yield satisfactory performance in the literature. 
Therefore, the conducted quantitative analysis in Section~\ref{Section2} motivates us to explore a deep learning-based approach for seizure prediction using iEEG data. The proposed approach in Section~\ref{Section3} has the desired advantages as described above.

\section{Methodology}
\label{Section3}

Convolutional Neural Networks (CNNs) have been proven to achieve astonishing results in different research areas such as face recognition \cite{Taigman}, object detection \cite{Erhan_2014_CVPR}, image classification \cite{Krizhevsky}, and speech recognition \cite{Graves}. In this study, we propose the use of CNNs for epileptic seizure prediction. 

We rely on CNNs to extract different representations of the iEEG data to use for iEEG classification. The proposed neural network architecture is unique as it allows us to recover both the local features (through smaller convolutions) and the high abstracted features (with larger convolutions). Preprocessing and preparing the data for the feature extraction and classification processes are explained below.

\subsection{iEEG Preprocessing}

The first step of building our pipeline system is to preprocess the time-series iEEG data by 1) reducing the data dimensionality, 2) dividing the data into smaller segments, and 3) encoding the data into an image-like format.

\subsubsection{Data Reduction} ~\\  
A key limitation in the seizure prediction problem is the large data size. The data used had a total of 5,612 iEEG clips (4827 interictal and 785 preictal); each clip is a 10 minutes duration, and given the high sampling rate of 400Hz, each iEEG clip had 240,000 data-points (10min$\times$60sec$\times$400Hz). The total size of the three patients' iEEG data was around 120GB. Reducing the data size can significantly reduce the required memory and computational resources.  

\begin{figure*}[!ht]
	\centering
	\includegraphics[width=1.00\textwidth, height=0.90\textwidth]{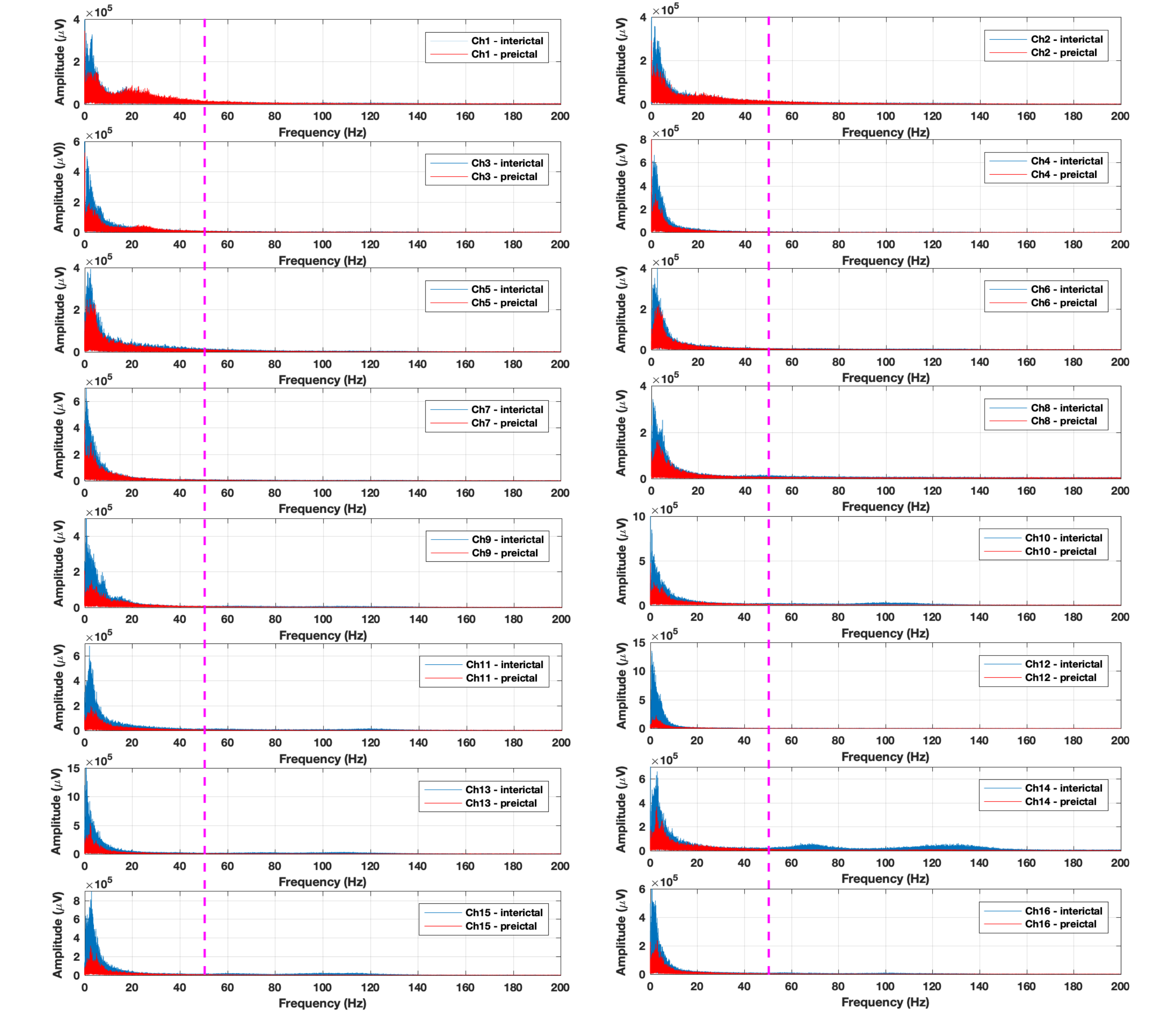}
	\caption{Frequency spectra of interictal (blue) and preictal (red) iEEG signals collected by the 16 implanted electrodes (channels) of the seizure advisory system in Patient~1. Ch1-Ch16 stand for Channels 1 to 16.}
	\label{Fig_iEEG_Spectrum}
\end{figure*}

As discussed in Section~\ref{Section2}, PCA is not suitable for dimensionality reduction of human iEEG data. We use a simple alternative for data reduction by resampling the iEEG clips at 100Hz rather than 400Hz. Figure~\ref{Fig_iEEG_Spectrum} shows an example of the frequency spectra of interictal and preictal iEEGs recorded by the 16-channel seizure advisory system in Patient~1. It can be noticed that, for both interictal and preictal iEEGs, a significant amount of iEEG information is concentrated in the low-frequency band ($<$50Hz), where there is no much relevant information in the high-frequency band ($>$50Hz). Following Nyquist rule, we thus resample the 10-minutes iEEG clips at 100Hz (2$\times$50Hz) instead of 400Hz. This reduces the dimensionality of the data by a factor of 4. The resulting number of data-points in an iEEG clip is thus 60,000 (10min$\times$60sec$\times$100Hz).

\begin{figure*}[!ht]
	\centering
	\includegraphics[width=1.00\textwidth]{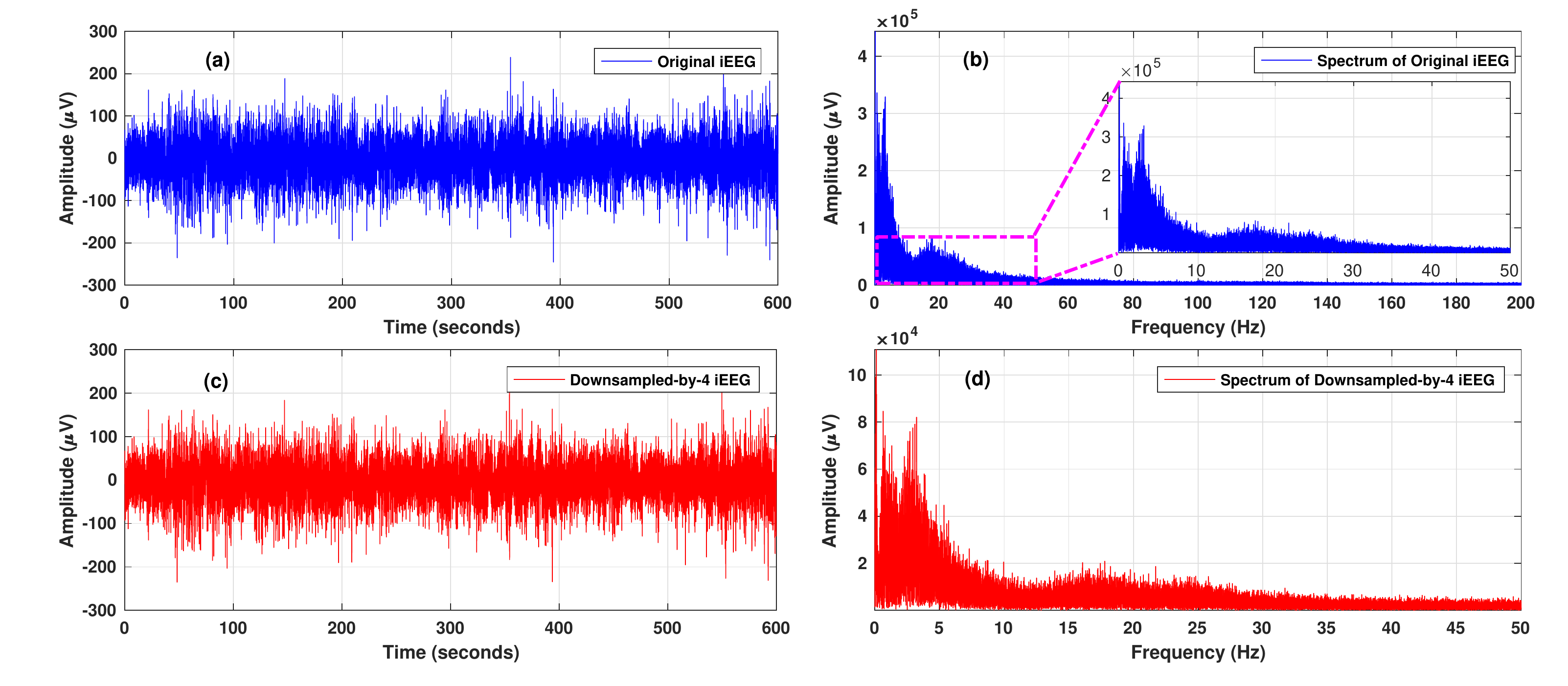}
	\caption{Time-series iEEG signals and their corresponding spectra: (a) and (b) original iEEG clip and its 200Hz frequency spectrum; (c) and (d) downsampled-by-4 iEEG clip and its 50Hz frequency spectrum.}
	\label{Fig_iEEG_Downsampling}
\end{figure*}

Figure~\ref{Fig_iEEG_Downsampling} demonstrates how the down-sampling process affects iEEG signals in both time and frequency domains. Figure~\ref{Fig_iEEG_Downsampling}(a) shows an example of time-series preictal iEEG signal and Fig.~\ref{Fig_iEEG_Downsampling}(b) depicts its spectrum along the entire frequency range of 0-200Hz. Figure~\ref{Fig_iEEG_Downsampling}(c) shows the down-sampled (by a factor of 4) version of the iEEG signal in Fig.~\ref{Fig_iEEG_Downsampling}(a). The frequency spectrum of the down-sampled signal is shown in Fig.~\ref{Fig_iEEG_Downsampling}(d).
It can be noticed in the figure that the power spectrum of the signal is concentrated in the frequency band of 0-50Hz. It is important to note that, in some cases, this down-sampling process may hurt the prediction performance, because some of the iEEG samples have energy beyond 50Hz.

\subsubsection{iEEG Segmentation} ~\\
Human surface and intracranial EEGs are usually non-stationary signals, \textit{i.e.}, their statistical characteristics change over time \cite{NonStationary1}. The main intuition behind iEEG segmentation is to divide a non-stationary iEEG signal into several pseudo-stationary segments as those are expected to have comparable features \cite{NonStationary2}. Another advantage of segmentation is the output large number of labeled iEEG samples needed for training deep CNNs to improve their performance. In this work, each down-sampled 10-minute iEEG clip is further split into 10 non-overlapping 1-minute iEEG segments; each has 6,000 data-points. This results in a tenfold increase in the total number of training (interictal and preictal) iEEG samples.

\subsubsection{Mapping Time-series iEEG Data into Images} 
\begin{figure}[!t]
	\centering
	\includegraphics[width=1.0\columnwidth, height=0.70\columnwidth]{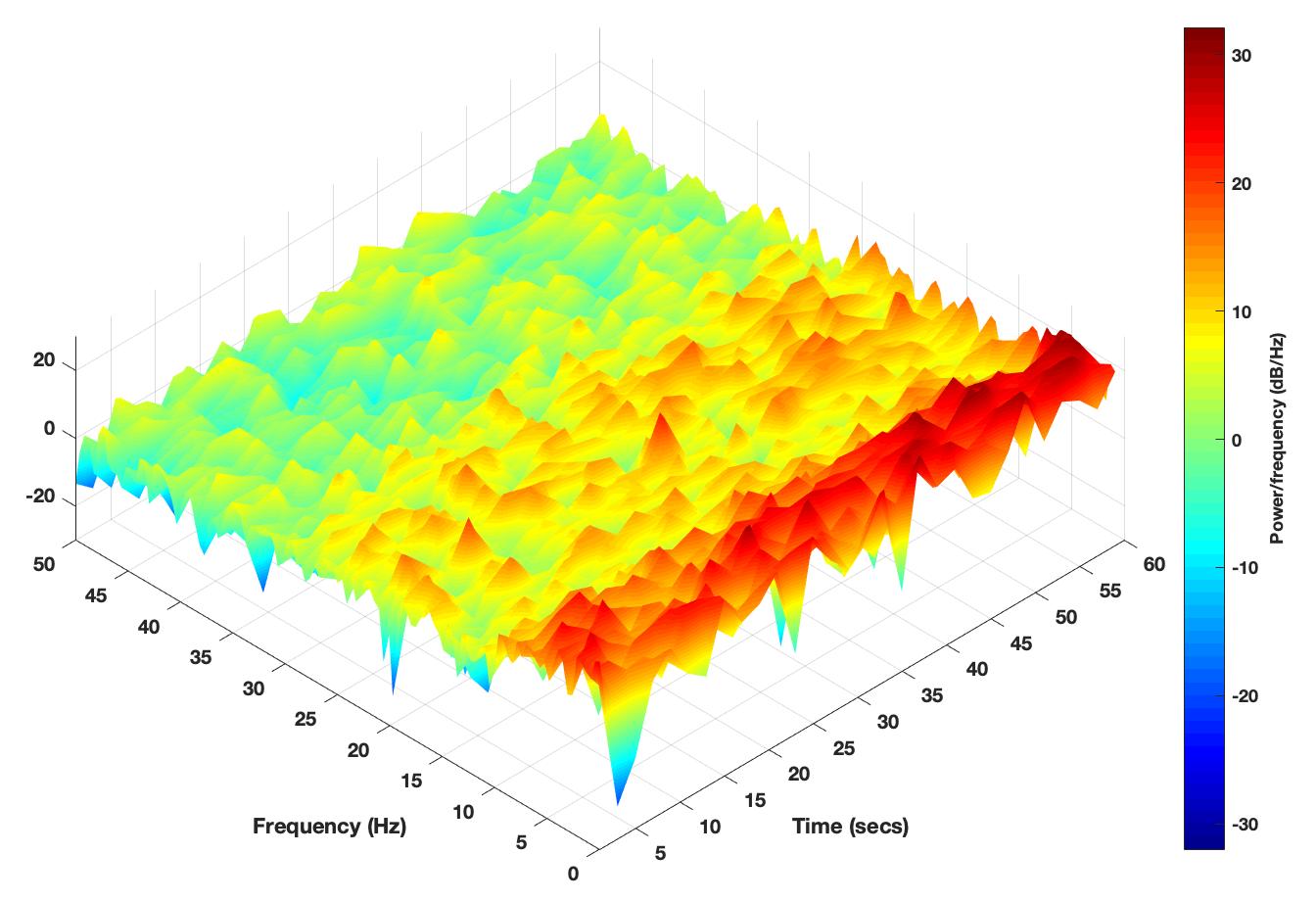}
	\caption{3D Spectrogram of a 1-minute preictal iEEG segment.}
	\label{Fig_Spectrogram_Jet}
\end{figure}

~\\
Since CNN has achieved great success in computer vision tasks, we are motivated to convert time-series iEEG data into an image-like format. We propose the use of Short-Time Fourier Transform (STFT) to obtain a two-dimensional representation of each iEEG segment. The (6,000 data-points) 1-minute iEEG segments are first partitioned into shorter chunks of equal length, and then the Fourier transform is computed for each individual chunk. This eventually reveals the changing power spectra as a function of time and frequency. Figure~\ref{Fig_Spectrogram_Jet} shows the three-dimensional spectrogram of a 1-minute preictal iEEG segment. It can be noticed that the iEEG signal power decays along the frequency axis starting with 30dB at 0-5Hz and till $-$30dB at 40-50Hz.       

After iEEG downsampling and segmentation, the resulting dimension of the data is 10$N$$\times$16$\times$6,000; where $N$ is the total number of 10-minute iEEG training clips, 10 is the number of iEEG segments per clip, 16 is the number of iEEG channels, and 6,000 is the length of each iEEG segment. For each iEEG segment, we compute the STFT for the 16 channels separately. The resulting dimension of the data is thus 10$N$$\times$16$\times$129$\times$26; where 129 and 26 are the numbers of frequencies and time chunks, respectively.

\begin{figure*}[!ht]
	\centering
	\includegraphics[width=1.00\textwidth]{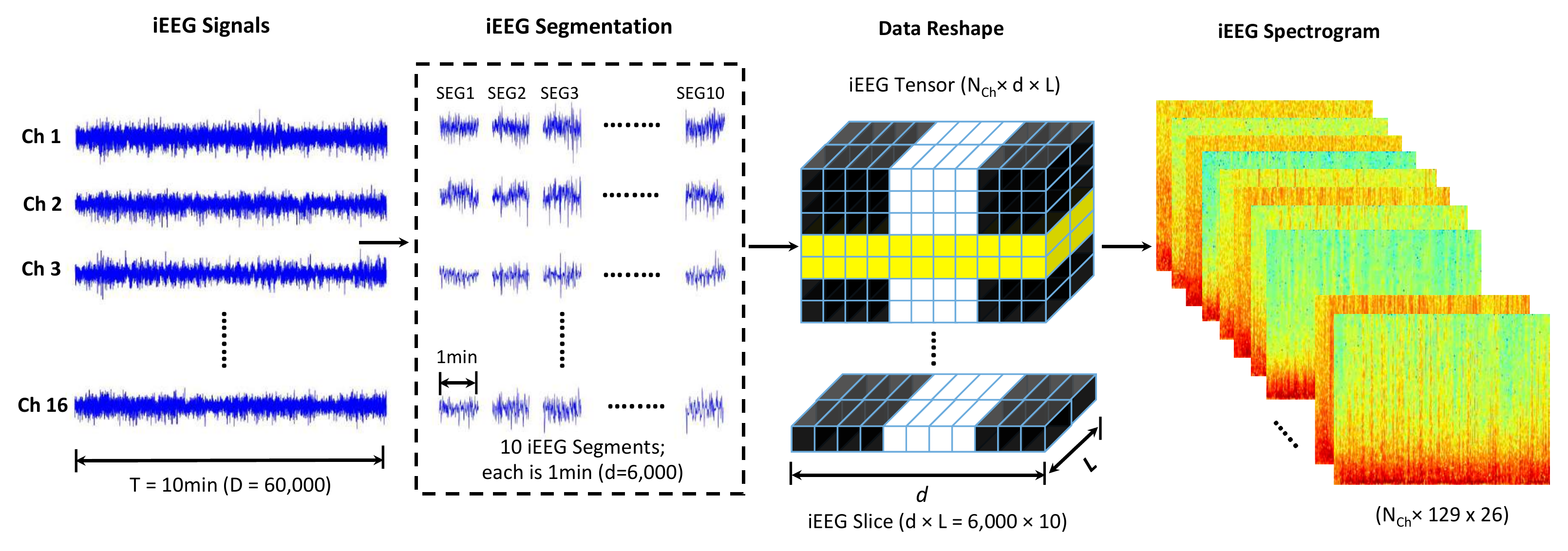}
	\caption{Schematic pipeline of the proposed iEEG data pre-processing approach for epileptic seizure prediction: SEG1, SEG2, $\cdots$, SEG10 are corresponding to the $1^{st}$, $2^{nd}$, and $10^{th}$ iEEG segments of each iEEG channel signal; $N_{Ch}$ is the total number of iEEG channels ($N_{Ch}$=16), $L$ is the number of segments per iEEG clip ($L$=10), and $d$ is the number of data-points in each iEEG segment ($d$=6,000).}
	\label{Fig_iEEG_Preprocessing}
\end{figure*}

Figure~\ref{Fig_iEEG_Preprocessing} illustrates all the transformations applied to the raw iEEG data before feeding it to the seizure prediction algorithm. The final output is STFTs of 1 minute duration and 50Hz bandwidth. To standardize the training data across all 16 channels, the mean and standard deviation were computed for each STFT image in every channel and then used to standardize the test set prior to their classification. 

\subsection{Automatic Feature Learning of iEEG.} ~\\

\begin{figure*}[!ht]
	\centering
	\includegraphics[width=1.00\textwidth
	]{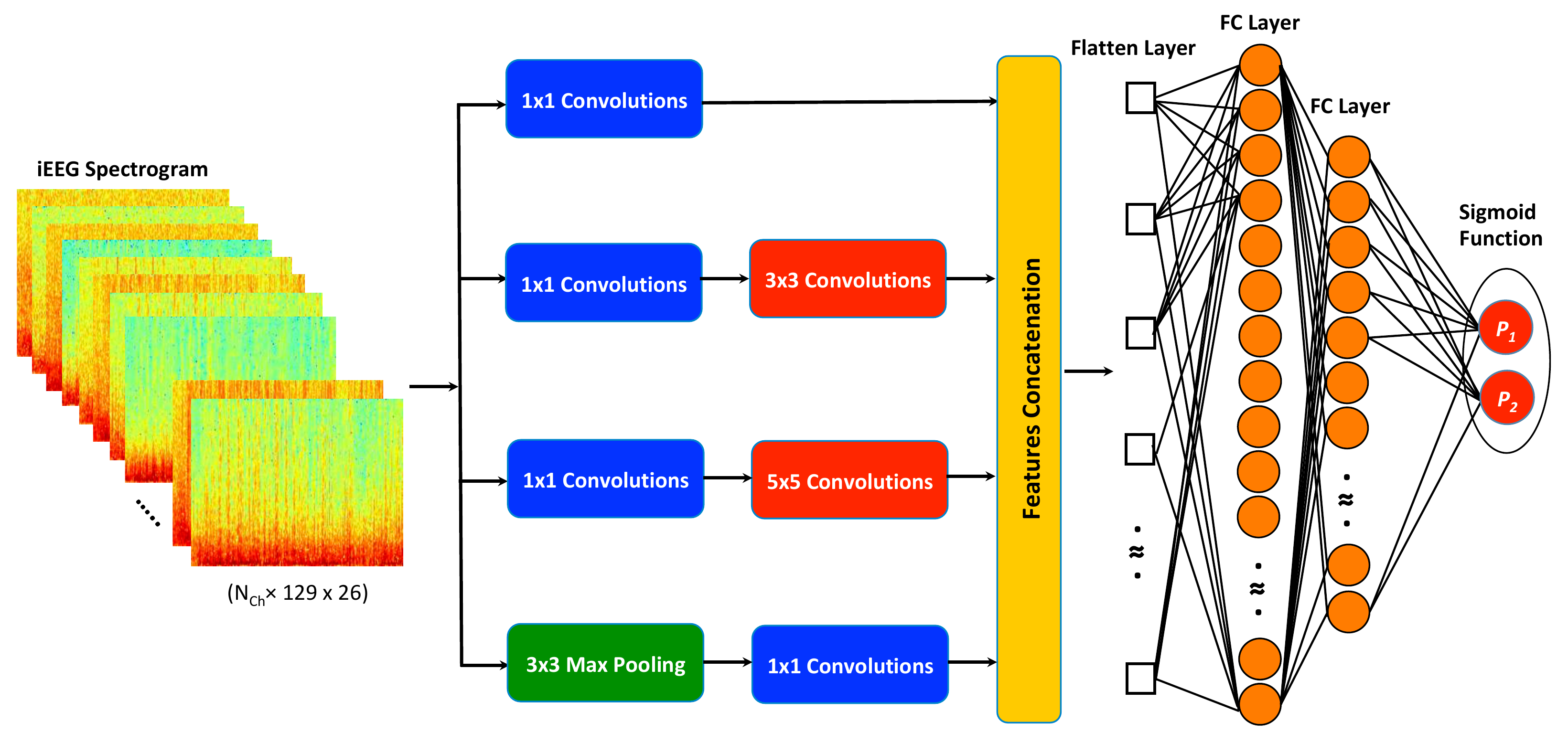}
	\caption{Schematic diagram of the proposed neural network architecture for epileptic seizure prediction: Each iEEG instance is of N$_{Ch}$$\times$129$\times$26 (N$_{Ch}$=16); Max Pooling stands for Maximum Pooling; FC layer stands for Fully Connected layer; $P_1$ and $P_2$ are the probabilities produced by the sigmoid function got class 1 and 2, respectively.}
	\label{Fig_NN_Model}
\end{figure*}

Figure~\ref{Fig_NN_Model} depicts the detailed architecture of the proposed CNN-based seizure prediction method. As explained in the previous subsection, the input data takes the shape of 10N$\times$16$\times$129$\times$26; where 10N is the total number of iEEG samples. Each sample comprises 16 STFT images; one for each iEEG channel. These images are first fed into a combination of CNNs of the same filter size (1$\times$1) with their output supplied to other CNNs of bigger filter size (3$\times$3 and 5$\times$5). Additionally, since pooling operations have been proven to play a key part in the promising convolutional networks, we add a parallel path in which maximum pooling and convolution operations have been adopted. Ultimately, the outputs of all parallel paths are concatenated into a single feature vector forming the input of the next few layers. The concatenated output is flattened and presented as an input to two Fully Connected (FC) layers. Finally, a Sigmoid function is used to compute the label probabilities and predictions \cite{Softmax}.

As for the proposed architecture, we were inspired by the successful CNN architectures in computer vision tasks. The goal here is to learn different feature maps and then combine them together so that the following FC layers can learn discriminative features from different scales simultaneously. The proposed CNN model was trained using 50,470 iEEG spectrograms (43,140 interictal and 7,330 preictal), and its seizure prediction performance was tested using 5,830 iEEG spectrograms (5,310 interictal and 520 preictal)\footnote[2]{Our experiments on the human iEEG-based seizure prediction using multi-scale CNNs were conducted with the open-source software of Keras using TensorFlow backend \cite{OurCodes}.}. To make a single prediction for each 10-minute iEEG clip, we take the maximum probability over the past consecutive ten 1-minute predictions. The arithmetic mean was also computed for every consecutive ten predictions, however, it achieved inferior prediction results than those of the maximum.    

From the network configuration perspective, our CNN model was trained by optimizing the ``binary cross-entropy'' cost function with ``Adam'' parameter update and a learning rate of 0.001. The 1$\times$1, 3$\times$3, and 5$\times$5 convolutional networks have 64 units each. The number of units for the two FC layers was set to 124 and 64, respectively. Our implementation was derived in Python using Keras with TensorFlow backend and the training took eight hours on an NVIDIA K40 GPU machine.

\subsection{Performance Evaluation.}

Metrics used to assess the prediction performance of the proposed method are sensitivity (SEN) and area under the ROC curve (AUC). To examine the generalizability of our seizure prediction algorithm over different subjects, we first evaluate the performance metrics for the three patients individually and then report the average performance.


\section{Results and Discussion}
\label{Section4}

In this section, we first assess the seizure prediction performance of the proposed algorithm and compare it to the state-of-the-art methods when tested on the same benchmark iEEG dataset \cite{cook2013prediction}. We also provide four intuitive reasons why current seizure prediction algorithms achieve limited performance.


\subsection{Seizure Prediction Results}

In this section, we test our seizure prediction algorithm and compare the results with those of four recent studies (reported in \cite{cook2013prediction}, \cite{karoly2016interictal}, \cite{karoly2017circadian}, and \cite{kiral2018epileptic}) that were tested on the same human iEEG dataset. In \cite{cook2013prediction}, Cook \textit{et al.} implanted the first-in-man Seizure Advisory System (SAS) in patients with refractory epilepsy. After SAS implantation, a seizure prediction algorithm was developed to distinguish time intervals of low, medium, and high likelihood of upcoming seizure attacks. Satisfactory seizure prediction results were achieved for most of the patients, proving that seizure prediction was possible. For example, the average seizure prediction sensitivity for all patients was around 61.20\%, while the three patients under study had inferior seizure prediction sensitivities with an average of 33.67\%.
The main reason behind such low seizure prediction performance was the temporal drift (variation) observed in the adopted time-dependent iEEG features (this study did not report what kind of iEEG features were extracted and what classifier was used to evaluate their effectiveness) \cite{cook2013prediction}.      

In \cite{karoly2016interictal}, Karoly \textit{et al.} developed a detection method to identify probable seizure activities. They used the preictal spike rate as an indicator that the brain is approaching a seizure. However, the spike rate was found to be an unreliable biomarker for all patients, as it was shown to increase before seizures for 9 out of 15 patients and decrease for the rest of the patients. The classification of iEEG clips was performed manually and the results showed an average seizure prediction sensitivity of 66.54\% for all patients and 43.34\% for the three patients under study.
In \cite{karoly2017circadian}, the same authors proposed a circadian seizure forecasting method to achieve robust prediction performance across all patients. With the help of the adopted circadian information, a significant improvement in the predictive power of seizure prediction methods was observed for most of the patients. The logistic regression classifier was used to assess the prediction performance of the proposed algorithm. Average seizure prediction sensitivities of 62.10\% and 52.67\% were achieved for all patients and for the three patients under study, respectively. In \cite{kiral2018epileptic}, Kiral-Kornek \textit{et al.} used deep learning to build an accurate seizure warning system that is patient-specific and can be adjusted to meet patients' needs. The deployed algorithm manifested notable seizure prediction results for all patients; yielding an average sensitivity of 69.00\% for the 15 patients and 77.36\% for the three patients whose data is studied in this work.

\begin{table*}[!t]
	\small
	\centering
	\caption{Seizure prediction sensitivity scores of the algorithms under study.}
	\label{Tab_Results_Sensitivity}
	\begin{tabular}{c c c c c c c}
		\hline
		\hline
		Subject & Cook \textit{et al.} \cite{cook2013prediction} & Karoly \textit{et al.} \cite{karoly2016interictal} & Karoly \textit{et al.} \cite{karoly2017circadian} & Kiral-Kornek \textit{et al.} \cite{kiral2018epileptic} &  Proposed Method &  Proposed Method\\
		
		& (2013) & (2016) & (2017) & (2018) &  (DS-by-4) &  (DS-by-2) \\
		\hline
		Patient 1 & 45.00 & 56.00 & 55.00 & 71·10 & \textbf{82.92} & 79.65 \\
		Patient 2 & 17.00 & 18.00 & 45.00 & 83·10 & 88.52 & \textbf{91.86} \\
		Patient 3 & 39.00 & 56.00 & 58.00 & 77·90 & 91.84 & \textbf{92.05} \\
		\hline
		Average & 33.67 & 43.34 & 52.67 & 77.36 & 87.76 & \textbf{87.85} \\
		\hline
		\hline
	\end{tabular} 
	\raggedright
	\justify 
	Patient~1, 2, and 3 in the Kaggle compeition dataset are the same as Patients~3, 9, and 11 in \cite{cook2013prediction}, \cite{karoly2016interictal}, \cite{karoly2017circadian}, and \cite{kiral2018epileptic}. DS = downsampled. DS-by-4 and DS-by-2 indicate the seizure prediction results using the downsampled-by-4 and downsampled-by-2 iEEG data, respectively.
\end{table*}

It is worth noting that the seizure prediction algorithms presented in \cite{cook2013prediction}, \cite{karoly2016interictal}, \cite{karoly2017circadian}, and \cite{kiral2018epileptic} could not achieve adequate prediction sensitivity for Patients~3, 9, and 11. These are the same as Patients~1, 2, and 3 in our study. The potential reasons for such poor performance are given in the following subsection. In our work, a multi-scale CNN architecture was specifically built to accurately recognize pre-seizure patterns in the preictal iEEG data taken from those three patients. Table~\ref{Tab_Results_Sensitivity} summarizes the seizure prediction results achieved by the proposed and the state-of-the-art methods. It can be noticed that, for the three patients, our algorithm yields superior seizure prediction sensitivity rates. It is worth mentioning that the seizure prediction performance is affected by the number of training iEEG samples for individual patients. For instance, Patient~3 has the largest number of training samples (see Table~\ref{Tab_iEEGDescription}), which in turn helps our CNN algorithm achieve a notably high seizure prediction sensitivity of 91.84\%. Nevertheless, a lower prediction sensitivity of 82.92\% was achieved for Patient~1 which had the least number of training samples.


As explained in Section~\ref{Section3}, all the iEEG signals are downsampled by a decimation factor of 4 for the purpose of speeding up the deployed deep learning algorithm. The question that may arise here: Does the downsampled-by-4 data closely represent the original data? 
Resampling the data at 100Hz means that all iEEG signals' content above 50Hz would be discarded. After careful examination of the frequency spectra of a wide range of interictal and preictal iEEG signals, we have found the following: (i) for most of the iEEG signals, the signal power dwells in the low-frequency band of 0-50Hz (as shown in Fig.~\ref{Fig_iEEG_Spectrum}), and (ii) for a few iEEG signals, a non-trivial amount of the signal power resides in higher frequency bands ($>$50Hz), implying that downsampling the data at 100Hz entails a considerable loss of information.  

In an attempt to retain as much iEEG information as possible, the original iEEG signals are downsampled by a factor of 2 (\textit{i.e.}, $f_s$=200Hz). The proposed seizure prediction algorithm is then tested on the data after downsampling it by a factor of 2. The achieved seizure prediction sensitivities are reported in Table~\ref{Tab_Results_Sensitivity}. The results demonstrate that downsampling the data at 200Hz helps preserve the iEEG signal information below 100Hz, and hence improve the overall seizure prediction sensitivity to 87.85\%.   



In addition, we compare our results to those of the winning solutions in the 2016 Kaggle seizure prediction challenge. The receiver operating characteristic (ROC) area under the curve (AUC) was the metric used for ranking various solutions. The winning team deployed 11 different classification models and more than 3000 hand-crafted iEEG features, and achieved an average AUC of 0.854 (see Table~\ref{Tab_Results_AUC}) \cite{WinningSolution, kuhlmann2018epilepsyecosystem}. It is, however, impractical to use such a computationally-intensive process for real-time applications. Our algorithm obtains a comparable but slightly lower seizure prediction AUC score of 0.840 on the iEEG data downsampled by 2. It is, however, much faster in obtaining the results and is thus more suitable for use in ambulatory and clinical applications. 
For faster iEEG processing (feature extraction and classification), the proposed algorithm is also tested on the downsampled by 4 iEEG data and an average AUC score of 0.787 is achieved. 


\begin{table}[!t]
	\small
	\centering
	\caption{AUC scores for the proposed method and Kaggle top finishing contestants on the public test set.}
	\label{Tab_Results_AUC}
	\scalebox{0.85}{
		\begin{tabular}{c c c}
			\hline
			\hline
			Top 10 Kaggle Scores (2018) \cite{kuhlmann2018epilepsyecosystem} & Proposed Method & Proposed Method \\
			min-max & (DS-by-4) & (DS-by-2) \\
			\hline
			0.820-0.854 & 0.787 & 0.840  \\
			\hline
			\hline
		\end{tabular}
	}
	\raggedright 
	\justify
	The first column shows the minimum and maximum AUC scores of the 10 top-scoring solutions of a post-competition study \cite{kuhlmann2018epilepsyecosystem}. The second and third columns show the AUC score of our multi-scale CNN model on the iEEG data downsampled by 4 and 2, respectively.
\end{table}


Table~\ref{Tab_Results_AUC_PerSubject} shows the AUC scores of our algorithm for individual patients when tested on iEEG data downsampled by 4 and 2. It can be seen that, for the downsampled-by-4 data, the AUC scores of 0.822 and 0.865 are achieved for Patients~2 and 3, respectively, while Patient~1 experiences a lower AUC score of 0.675. Using the downsampled-by-2 data helped improve the AUC scores for all patients. For example, a remarkable AUC score of 0.938 was achieved for Patient~3 (as this patient had the largest number of interictal and preictal iEEG samples). Patient~2 had a little lower AUC score of 0.891 and Patient~1 had the least AUC score of 0.692. The average AUC score produced by our CNN algorithm for all patients is 0.840.


\begin{table}[!t]
	\small
	\centering
	\caption{Per-subject AUC scores of the proposed method.}
	\label{Tab_Results_AUC_PerSubject}
	\begin{tabular}{c c c}
		\hline
		\hline
		Subject & Proposed Method &  Proposed Method \\
		& (DS-by-4) & (DS-by-2) \\
		\hline
		Patient 1 & 0.675 & 0.692 \\
		Patient 2 & 0.822  & 0.891 \\
		Patient 3 & 0.865  & 0.938 \\
		\hline
		Average & 0.787 & 0.840 \\
		\hline
		\hline
	\end{tabular}
\end{table}

\subsection{Possible Reasons for Limited Performance}

Seizure prediction algorithms examined on the 2016 Kaggle/Melbourne University iEEG dataset provide limited performance. The top contestant, for example, achieved the highest AUC of 0.854 \cite{kuhlmann2018epilepsyecosystem, WinningSolution}. Herein, we list some potential reasons why existing machine learning and deep learning algorithms could not achieve any better performance.


\subsubsection{iEEG Data Corruption}
\begin{figure}[!t]
	\centering
	\includegraphics[width=1.0\columnwidth, height=0.75\columnwidth]{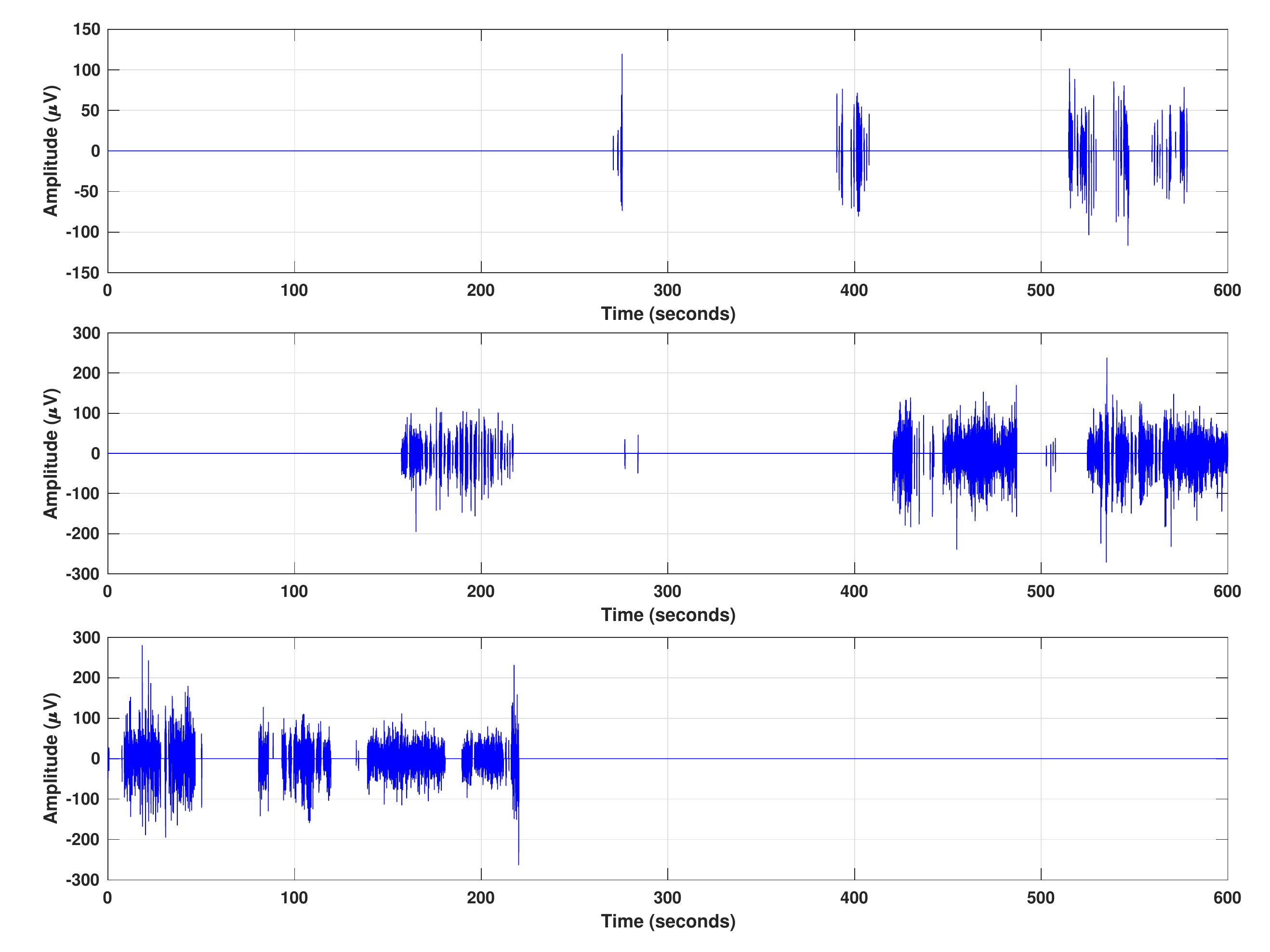}
	\caption{Examples of corrupted iEEG clips for Patient~1.}
	\label{Fig_CorruptediEEG}
\end{figure}

~\\
After a thorough inspection and data visualization of the iEEG data under study, we realized that many of the 10-minutes iEEG clips contain ``data drop-out''. This is when the implanted device failed to communicate with the storage device for several possible reasons. This data drop-out corresponds to iEEG signal values of zeros across all channels at a given time sample. A handful of 10-minute clips comprise 100\% data drop-out and cannot be classified. Other clips, however, are partially corrupted and contain different percentages of data drop-out. Fig.~\ref{Fig_CorruptediEEG} displays various examples of iEEG signals comprising data drop-outs at different time samples. It is worth noting that, if for any reason, a dataset includes corrupted instances, standard machine learning algorithms can break down.  


\subsubsection{iEEG Data Mismatch}
\begin{figure}[!t]
	\centering
	\includegraphics[width=1.0\columnwidth, height=0.82\columnwidth]{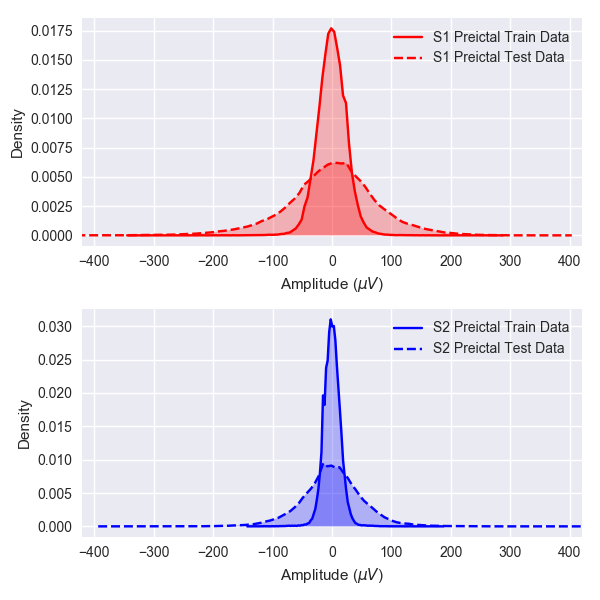}
	\caption{Data mismatch in Patient~1's preictal iEEG sensor data.}
	\label{Fig_DataMismatch}
\end{figure}

~\\
The terms ``data mismatch'', ``concept drift'', and ``covariate shift'' have been used to refer to the situation where data characteristics (distribution) change over time \cite{vzliobaite2016overview}. Often, the distribution of a particular data class (\textit{e.g.}, interictal or preictal) is assumed to not change over time, implying that the distribution of the historical data is just the same as the distribution of the new data. This holds true for many machine learning problems, but not for all problems. In some cases, the characteristics of the data vary over time, and hence the predictive models trained on historical data are no longer valid for making predictions on new unseen data. After attentive screening to the characteristics of the iEEG data under study, we realized that, for each individual patient, the distribution of either interictal or preictal iEEG data is different between the training set and testing set. A potential reason for such a data shift is that the testing data was recorded a long time after recording the training data. During this time, the patient may be positively or negatively influenced by the medications. Fig.~\ref{Fig_DataMismatch} depicts how the preictal iEEG data distributions for Patient~1 have changed over time. The top plot explains how the density of preictal iEEG signals recorded by Channel~1 differs between training and testing sets. Similarly, the bottom plot demonstrates the variations in preictal iEEG density for Channel~2.          


\subsubsection{Outliers in iEEG Data}
\begin{figure}[!t]
	\centering
	\includegraphics[width=1.0\columnwidth, height=0.78\columnwidth]{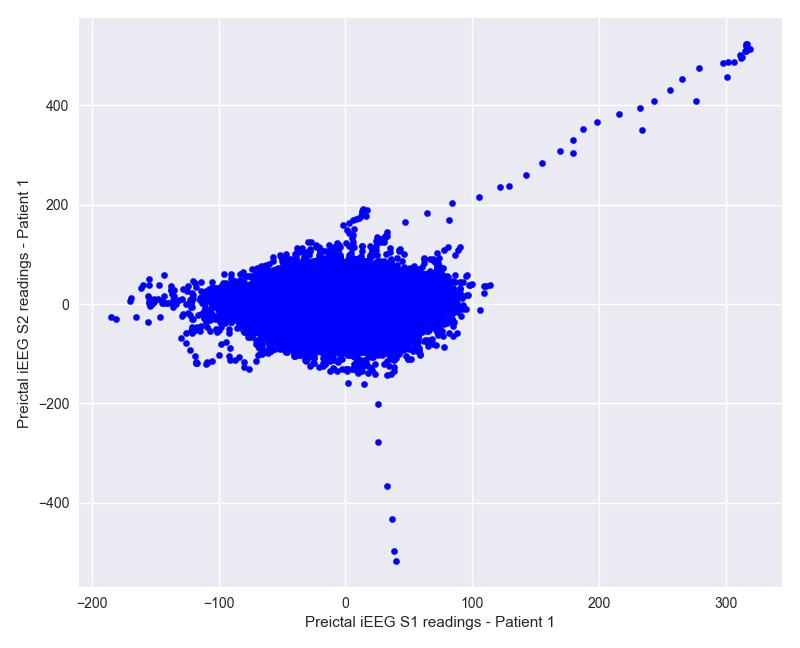}
	\caption{Scatter plot to identify outliers in preictal iEEG data of Patient~1: Preictal S1 readings vs. Preictal S2 readings.}
	\label{Fig_DataOutliers}
\end{figure}

~\\
Since iEEGs are recorded with electrodes attached directly to the surface of the brain, they are deemed to be free of body movement artifacts (\textit{e.g.}, muscle activities and eye blinking) and their signal-to-noise ratios are expected to be much higher than those of surface EEGs \cite{Nunez2010}. However, there exist some inevitable sources of artifacts (\textit{e.g.}, static magnetic field and environmental noise) that interfere with the iEEG signals producing undesired outliers. The scatter plot in Fig.~\ref{Fig_DataOutliers} shows an example of ``data outliers'' which were detected in the S1 and S2 readings for some preictal iEEG clips taken from Patient~1. Other scatter plots (omitted for the lack of space) demonstrate that, for the same iEEG clips, the remaining 14 sensor readings also include similar outliers. The presence of outliers often has a negative influence on the extracted features and certainly on the predictive power of the fitted models. If the outliers are removed from the fitting process, then the resulting fit will maintain subtle performance every time it is tested on new unseen data.


\subsubsection{Imbalanced Class Distribution}~\\
The seizure prediction dataset under study has 5,047 iEEG samples. A total of 4,314 belongs to Class-1 (interictal iEEG) and the remaining 733 samples belong to Class-2 (preictal iEEG). This represents a binary classification problem with imbalanced class distribution, where the ratio of Class-1 to Class-2 samples is 4,314:733, \textit{i.e.}, 5.885:1. With such an imbalanced dataset, the classification rules that predict the minority class tend to be more flimsy than those that predict the majority class; therefore, test samples belonging to the minority class are misclassified more often than those belonging to the majority class \cite{sun2006boosting}. Thus, one of the possible reasons for limited seizure prediction performance is that the deployed machine learning algorithms do not take into consideration the interictal and preictal class distributions. They pay less attention to the minority class of preictal iEEGs, which is of crucial importance for forecasting impending seizures.

There exist many approaches that can tackle classification problems having imbalanced data. One of the effective procedures is to generate synthetic samples from the under-represented class. In this study, we have first used the Synthetic Minority Over-sampling Technique (SMOTE) \cite{chawla2002smote} to generate more samples from the minority class (preictal iEEG). However, a serious degradation in the seizure prediction performance was noticed (as the trained CNN model was overfitting). As an alternative solution, we employed the class weights in training the proposed CNN model in order to force the model to treat every sample of preictal iEEG as $\sim$6 samples of interictal iEEG. This helped achieve the remarkable seizure prediction performance reported in Tables~\ref{Tab_Results_Sensitivity}, \ref{Tab_Results_AUC}, and \ref{Tab_Results_AUC_PerSubject}.   




\section{Conclusion}
\label{Section5}

Big data is crucial for reliable seizure prediction. The Kaggle/Melbourne University Seizure Prediction Competition provides a big dataset. It includes a long-term human intracranial EEG data recorded continuously from three (drug-resistant) epileptic patients for over 373-559 days. Leveraging this dataset properly can lead to substantial improvements in the field of epileptic seizure prediction. 


We have therefore done an extensive analysis of the iEEG data, and from which we have provided,
for the first time, a detailed quantitative analysis of the human iEEG data during the preictal and interictal brain states. This analysis reveals why hand-crafted iEEG features that have been successfully used in seizure detection are not the best candidates for robust prediction of epileptic seizures. We then proposed a convolution neural network (CNN) model for accurate prediction of epileptic seizures. The data was first divided into smaller segments on which we applied STFT. This resulted in a much larger number of interictal and preictal iEEG samples, which is important in deep learning in achieving better seizure prediction performance. 
The resulting time-series segments were converted into image-like formats to enable its use as inputs to our CNN model. The CNN model we proposed has an efficient multi-scale CNN architecture that automatically learns distinctive iEEG features from different spatial scales contemporaneously.
Unlike previous seizure prediction algorithms that used hand-crafted features, the proposed algorithm can yield reliable performances across different patients, achieving a superior average prediction sensitivity performance as 87.85\%.


Though training the proposed CNN model takes 6-8 hours, the trained model takes less than a second to test new unseen data. Compared to the Kaggle leaderboard winning solution that uses over 3,000 custom features, the proposed algorithm achieves comparable performances in a much more computationally-efficient fashion.





\bibliographystyle{IEEEtran}
\bibliography{References}

\end{document}